\documentclass[conf, hidelinks]{IEEEtran}
\usepackage[utf8]{inputenc}

\usepackage{graphicx}
\usepackage{bm}
\usepackage{siunitx}
\usepackage{longtable,tabularx}

\usepackage{color}
\usepackage[capitalise]{cleveref}

\usepackage{algpseudocode}
\usepackage{algorithm}

\usepackage{caption}
\usepackage{subcaption}
\usepackage{textcomp}
\usepackage{hyperref}
\usepackage{xspace}
\newcommand{\blue}[1]{{\color{blue}#1}}

\newcommand{\linkToPdf}[1]{\href{#1}{\blue{(pdf)}}}
\newcommand{\linkToPpt}[1]{\href{#1}{\blue{(ppt)}}}
\newcommand{\linkToCode}[1]{\href{#1}{\blue{(code)}}}
\newcommand{\linkToWeb}[1]{\href{#1}{\blue{(web)}}}
\newcommand{\linkToVideo}[1]{\href{#1}{\blue{(video)}}}
\newcommand{\linkToMedia}[1]{\href{#1}{\blue{(media)}}}
\newcommand{\award}[1]{\xspace} %

\newcommand{\rulesep}{\unskip\ \vrule\ }

\title{Vision-Based Terrain Relative Navigation \\ on High-Altitude Balloon and Sub-Orbital Rocket}

\author{Dominic R. Maggio\footnote[1]{S.M. Candidate, Aeronautics and Astronautics, MIT; Draper Scholar w/ Perception and Embedded Machine Learning Group, Draper}}
\affil{Massachusetts Institute of Technology, Cambridge, MA, 02139, USA}
\author{Courtney Mario$^\dagger$, Brett Streetman\footnote[3]{Distinguished Member of the Technical Staff, Autonomy and Real-time Planning}, 
 Ted J. Steiner\footnote[2]{Principal Member of the Technical Staff, Perception and Embedded Machine Learning Group, Draper}}
\affil{The Charles Stark Draper Laboratory, Inc., Cambridge, MA, 02139, USA}
\author{Luca Carlone\footnote[4]{Associate Professor, Aeronautics and Astronautics, MIT}}
\affil{Massachusetts Institute of Technology, Cambridge, MA, 02139, USA}

\begin{document}

\maketitle

\begin{abstract}
    We present an experimental analysis on the use of a camera-based approach for high-altitude navigation by associating mapped landmarks from a satellite
    image database to camera images, and by leveraging inertial sensors between camera frames. 
    We evaluate performance of both a sideways-tilted and downward-facing camera on data collected from a World View
    Enterprises high-altitude balloon with
    data beginning at an altitude of 33 km and descending to near ground level (4.5 km) with 1.5 hours of flight time. 
    We demonstrate less than 290 meters of 
    average position error over a trajectory of more than 150 kilometers. 
    In addition to 
    showing performance across a range of altitudes, we also demonstrate the 
    robustness of the Terrain Relative Navigation (TRN) method to rapid rotations of the balloon, in some cases exceeding $20^\circ$
    per second, and to camera obstructions caused by both cloud coverage and cords swaying underneath the balloon. 
    Additionally, we evaluate performance on 
    data collected by two cameras inside the capsule of Blue Origin's New Shepard rocket on payload flight NS-23, traveling at speeds up to 880 km/h, %
    and demonstrate less than 55 meters of average position error.
    \end{abstract} %

\section{Nomenclature}

{\renewcommand\arraystretch{1.0}
\noindent\begin{longtable*}{@{}l @{\quad=\quad} l@{}}
$altitude$ & height above WGS84 ellipsoid \\
$ecef$  & Earth-centered, Earth-fixed coordinate system\\
$ENU$  & East, North, Up coordinate system\\
$\Delta t_{imu}$  & time between IMU measurements, s\\
$\bm{\theta}$ &    gyroscope measurements $[\theta_x, \theta_y, \theta_z]^T$, $\frac{rad}{s}$ \\
$\omega$& $\| \bm{\theta} \| $ \\
$q_{1}^{2}$ & unit quaternion describing the rotation from frame 1 to frame 2 \\
$P$ & camera projection matrix \\
$\pi_{WGS84}$ & projection of a pixel coordinate to a 3D point on the surface of the WGS84 model \\
$\alpha_{max}$ & max acceptable angle between camera boresight and normal of a landmark \\
$\delta x$ & amount to shift a point by in pixel space \\
$surface\_normal()$ & function that finds normal vector at a point on the WGS84 model \\
$angle\_between()$ & function that finds the angle between a camera boresight and a vector \\
\end{longtable*}} \clearpage

\section{Introduction}
\label{sec:intro}

Terrain Relative Navigation (TRN) is a method for absolute pose estimation in a GPS-denied environment using a prior map of the environment and onboard sensors such as a camera. TRN 
is commonly desired %
for applications requiring accurate pose estimation, such as planetary landings and airdrops, where GPS 
is either unavailable or cannot be relied upon. Due to the high altitude of planetary TRN missions, acquiring non-simulation test data oftentimes proves difficult, 
and thus many datasets used to test TRN systems are from lower altitudes than what the system would actually be used at during a mission. Additionally, 
for vision-based TRN systems, the large distance between the camera and features on the ground can make position changes of the camera 
difficult to accurately observe due to the high ratio of meters per pixel in the image plane.

\begin{figure}[hbt!]
    \centering
    \begin{subfigure}[t]{0.47\textwidth}
        \centering
        \includegraphics[width=.5\textwidth]{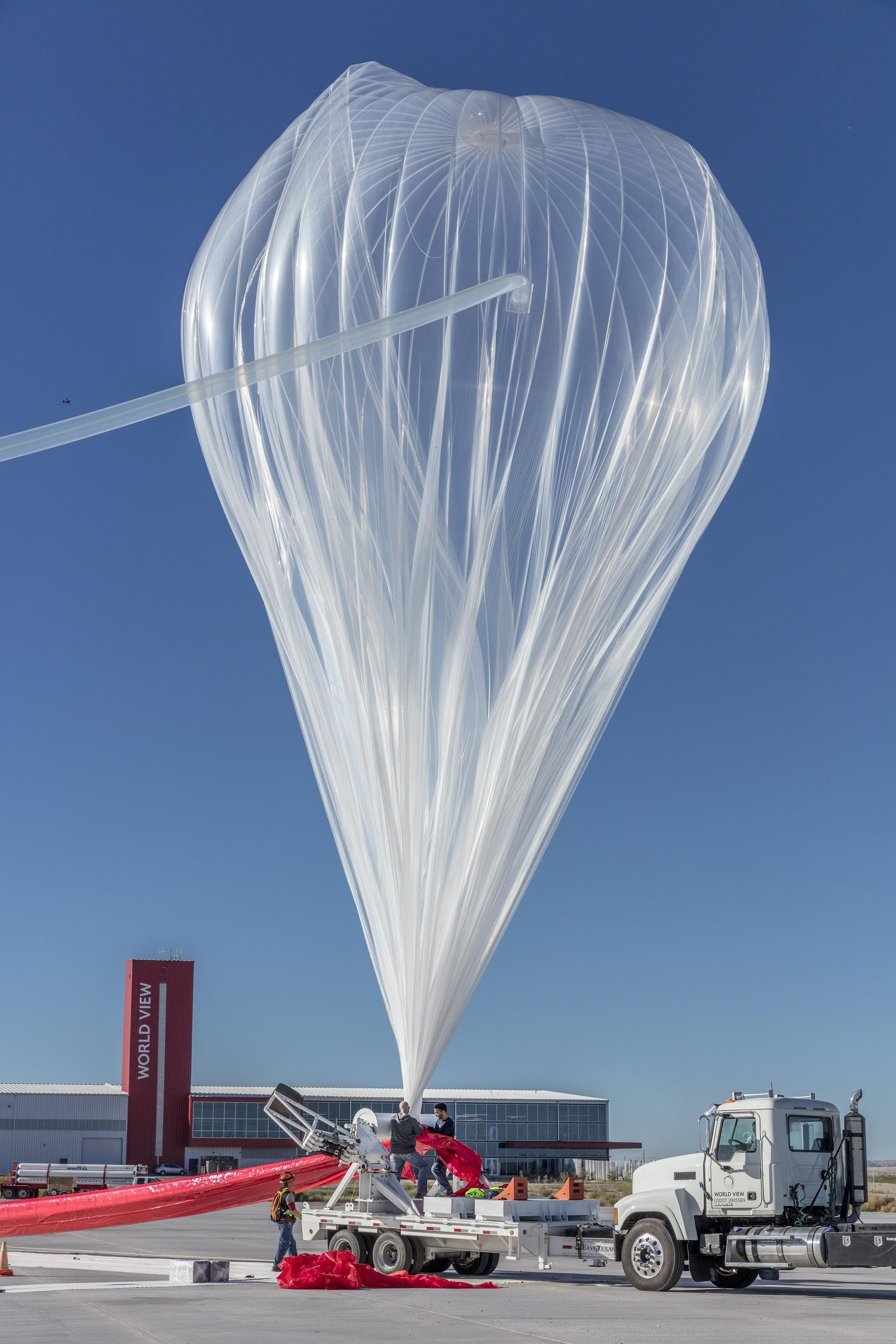}
        \caption{Release of high-altitude balloon for data collection. \\\ Image: courtesy of World View\textregistered  Enterprises}
        \label{fig:balloon_launch}
        \end{subfigure}
    \hfil
    \begin{subfigure}[t]{0.418\textwidth}
        \centering
        \includegraphics[width=.50\textwidth]{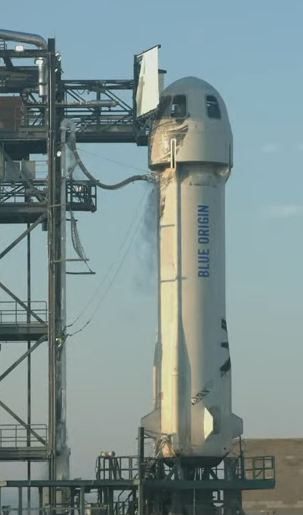}
        \caption{Blue Origin's New Shepard rocket carrying Draper experimental payload in the capsule. Image: courtesy of Blue Origin}
        \label{fig:rocket_all}
    \end{subfigure}
    \caption{Data collection platforms used for experimental analysis.}
\end{figure}

\begin{figure}[hbt!]
    \centering
    \includegraphics[width=1.0\textwidth]{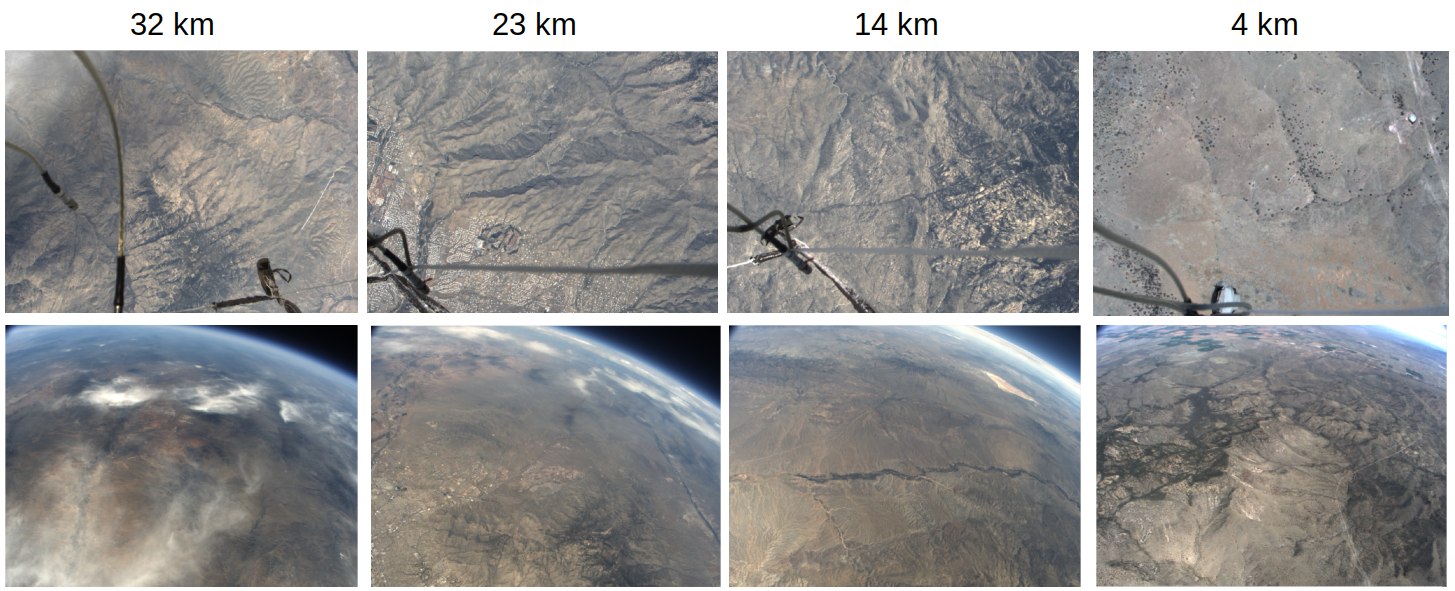}
    \caption{Example of images collected at different altitudes (32, 23, 14, and 4 km) from the balloon dataset with the downward-facing camera (top)
    and sideways-facing camera (bottom).}
    \label{fig:overview}
\end{figure}

\begin{figure}[hbt!]
    \centering
    \begin{subfigure}[t]{0.24\textwidth}
        \centering
        \includegraphics[width=\textwidth]{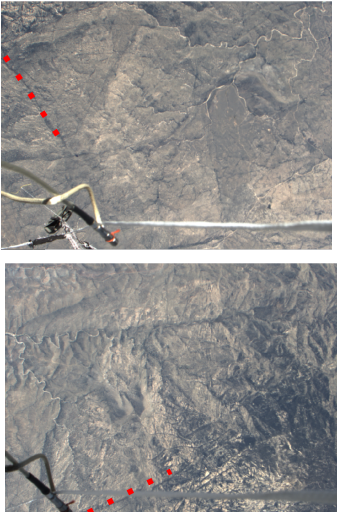}
        \caption{Rapid rotations, here over $90^\circ$ in 4 seconds. Red dots show ground reference points between top image and bottom image.}
        \label{fig:challenges_a}
    \end{subfigure}
    \rulesep
    \begin{subfigure}[t]{0.24\textwidth}
        \centering
        \includegraphics[width=\textwidth]{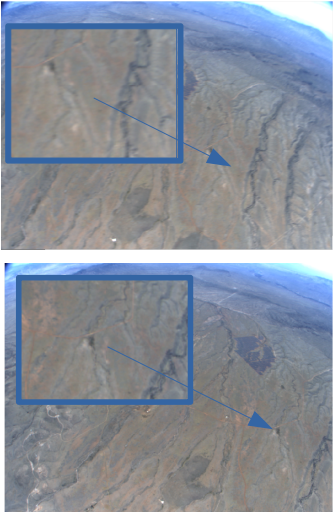}
        \caption{Image blur (top) due to rapid motion compared to crisp image (bottom).}
        \label{fig:challenges_b}
    \end{subfigure}
    \rulesep
    \begin{subfigure}[t]{0.24\textwidth}
        \centering
        \includegraphics[width=\textwidth]{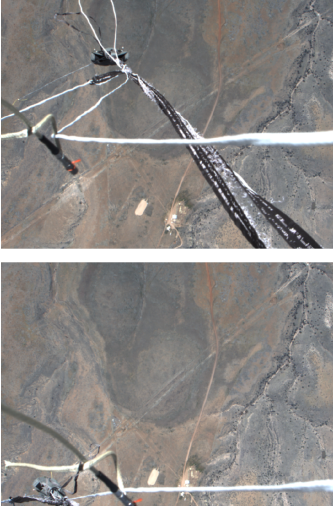}
        \caption{Moving cords in the image. Top and bottom images showing example range of cord motion.}
        \label{fig:challenges_c}
    \end{subfigure}
    \rulesep
    \begin{subfigure}[t]{0.24\textwidth}
        \centering
        \includegraphics[width=\textwidth]{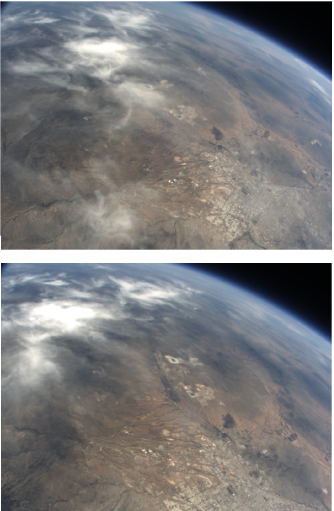}
        \caption{images partially occluded by clouds}
        \label{fig:challenges_d}
    \end{subfigure}
    \caption{Different types of TRN challenges in the balloon dataset.}
\end{figure}

This paper presents an experimental analysis on performing TRN using a camera-based approach aided by a gyroscope for high-altitude navigation by associating mapped landmarks from satellite
imagery to camera images. We evaluate performance of both a sideways-tilted and downward-facing camera on data collected from a World View Enterprises high-altitude balloon (\cref{fig:balloon_launch}) 
with
data beginning at an altitude of 33 km and descending to ground level with almost 1.5 hours of flight time (\cref{fig:overview}) and on data collected at speeds up to  
880 km/h (550 mph) from two sideways-tilted cameras mounted inside the capsule of Blue Origin's New Shepard rocket (\cref{fig:rocket_all}), during payload mission NS-23. We also demonstrate the 
robustness of the TRN system to rapid motions of the balloon which causes fast attitude changes (\cref{fig:challenges_a})
and can cause image blur (\cref{fig:challenges_b}). Additionally, we demonstrate performance in the presence of dynamic camera obstructions 
 caused by cords dangling below the balloon (\cref{fig:challenges_c}), and clouds obstructing sections of 
the image (\cref{fig:challenges_d}). 

\begin{figure}[hbt!]
    \centering
    \includegraphics[width=0.5\textwidth]{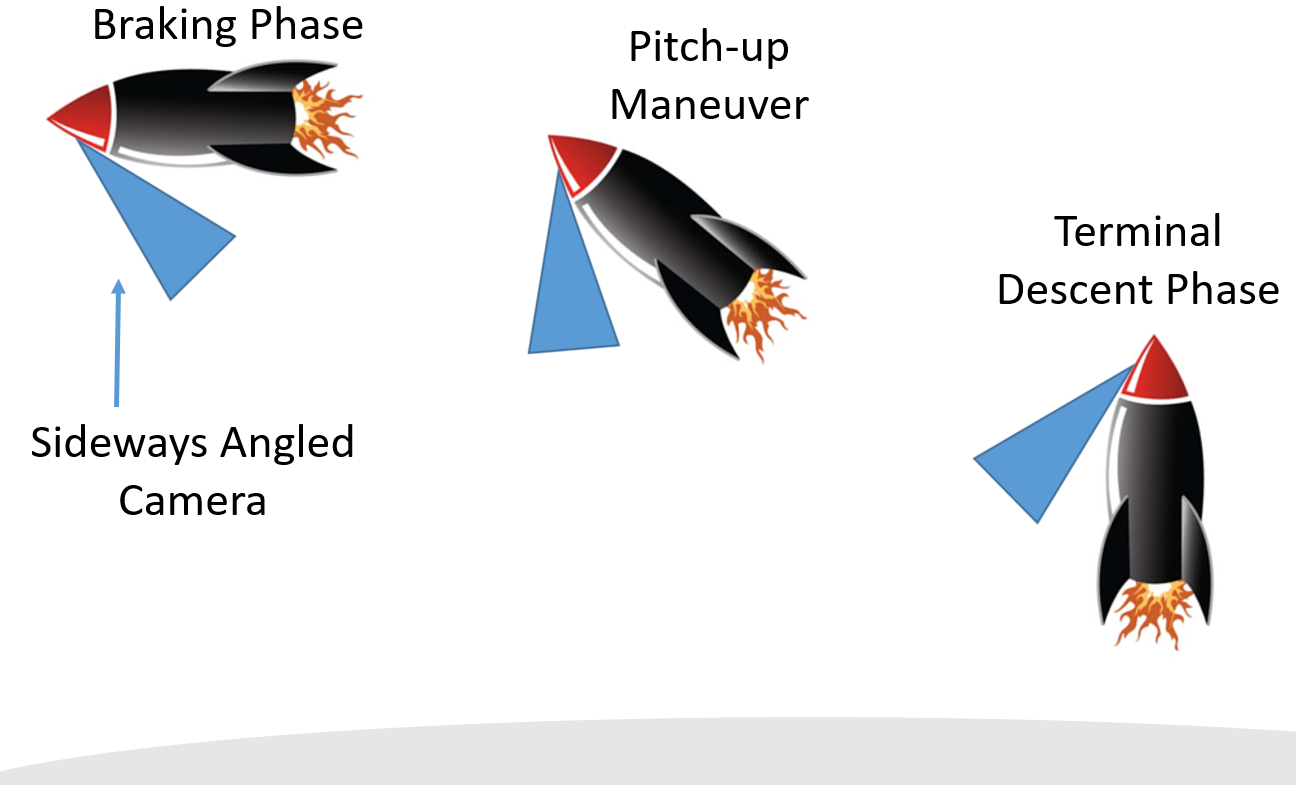}
    \caption{Demonstration of a sideways-angled camera viewing the terrain and being used 
    during the braking phase, pitch-up maneuver, and terminal descent phase.}
    \label{fig:landing}
\end{figure}

Sideways-angled cameras are a common choice for TRN applications when mounting a downward camera is either infeasible due to vehicle constraints or 
would be occluded by exhaust from an engine on vehicles such as a lander or a rocket. Additionally, for 
planetary landings, a sideways-angled camera allows for a single camera to be used 
during both the braking phase when the side of the lander faces the surface and during the 
final descent phase when the bottom of the lander faces the surface (\cref{fig:landing}). We thus use both a 
sideways-angled camera and downward-facing camera during our high-altitude balloon flight 
to separately evaluate the performance of TRN using a camera from each orientation.

We use Draper's Image-Based Absolute Localization (IBAL) \cite{Denver17aiaa-airdrop} software for our analysis. 
While our dataset has images at a rate of 20Hz, we subsample images by a factor of 10 and hence post-process images at 2Hz in real-time.
IBAL could additionally be combined with a nonlinear estimator such as an Extended Kalman Filter (EKF) or a fixed-lag smoother through either a loosely coupled approach using IBAL's pose estimate or a tightly-coupled approach using landmark matches~\cite{Forster17tro}. 
Since the quality of the feature matches generated by IBAL would affect all these methods, here we limit ourselves to evaluating IBAL as an independent system and also analyze the quality of the
feature matches. At the same time, we investigate the impact of using a gyroscope in conjunction with IBAL to aid with the challenges of our balloon dataset and show the advantage that 
even a simple sensor fusion method can provide. 
Finally, we extend IBAL to incorporate methods to 
efficiently process images when a camera views above the horizon.

\newpage

\section {Related Work}

We present an overview of existing Terrain Relative Navigation approaches and experiments, noting that our primary contribution are two experiments that
allow us to perform indepth analysis of vision-based terrain relative navigation on challenging high-altitude data and on data from a high speed vehicle. 
TRN methods primarily use either cameras, radar, or lidar as an exteroceptive sensor. The majority of 
early TRN methods such as the Mars Science Laboratory \cite{Katake10-landingNav} and NASA's ALHAT Project (\cite{Brady11gnc-alhat}, \cite{Amzajerdian12ac-lidarTRN}) 
use radar or lidar. However, due to the high power 
and weight budget of radar and lidar, cameras have been motivated as an active area of exploration for more recent TRN systems.

The seminal work of Mourikis \textit{et al.} \cite{Mourikis09tro-EdlSoundingRocket} describes a visual-inertial navigation method for 
Entry, Descent, and Landing (EDL) using an Extended Kalman Filter (EKF) with matched landmarks and
tracked feature points in an image. They use inertial navigation results from their entire sounding rocket launch with an apogee of 123 km, and leverage visual methods after the vehicle reaches altitudes below 3800m. Johnson and Montgomery~\cite{Johnson09ac-trnReview}
present a survey of TRN methods that use either image or lidar to detect the location of known landmarks.

Singh and Lim~\cite{Singh12aiaa-trnEKF} 
demonstrate a visual TRN approach leverging an EKF for lunar navigation using known crater locations as landmarks. Recently, Downes \textit{et al.} \cite{Downes20aiaa-lunarTRN} 
present a deep learning method for lunar crater detection to improve TRN landmark tracking. 
The Lander Vision System (LVS) \cite{Johnson17-lvs} used for the Mars 2020 mission uses vision-based landmark matching starting at an altitude of 4200m above the 
martian surface with the objective of achieving less than 40m error with respect to the landing site. 
Our analysis focuses on higher altitudes and on a larger span on altitudes (4.5 km to 33 km for the 
balloon dataset).

Dever \textit{et al.}\cite{Denver17aiaa-airdrop} demonstrate visual navigation for guided parachute airdrops using IBAL and a 
Multi-State Constraint Kalman Filter (MSCKF). Additionally, the work incorporates
a lost robot approach to recover from a diverged pose estimate and to initialize the system if the pose is unknown. 
Steffes \textit{et al.} \cite{Steffes19aiaa-trnEDL} present a theoretical analysis of three types of visual terrain navigation 
approaches, namely template matching, SIFT \cite{lowe2004ijcv-distinctive} descriptor matching, and crater matching. 
The  work of Lorenz \textit{et al.} \cite{Lorenz17ac-osirisrex} demonstrates 
vision-based terrain relative navigation for a touch and go landing on an asteroid for the OSIRIS-REx mission. Due to extreme computation limits,
they used a maximum of five
 manually selected mapped template features per frame. Mario \text{et al.} \cite{Mario22psj-osirisRexTesting} provide additional discussion on ground tests 
 used to prepare the TRN system for the OSIRIS-REx mission. Our balloon dataset has much faster rotional motion than 
 what was present during the OSIRIS-REx mission along with camera obstructions.

Steiner \textit{et al.} \cite{Steiner15ac-landmarkSelection} present a utility-based approach for optimal landmark selection and demonstrates performance 
on a rocket testbed flight up to 500m. As shadows and variable lighting conditions are a well known challenge for TRN, 
Smith \textit{et al.} \cite{Smith22aiaa-blenderTRN} demonstrates the ability to use Blender to enhance a satellite database for different lighting conditions. %

\section{Data Collection}

The collection of both datasets used in this paper was supported by the NASA Flight Opportunities Program. The high-altitude balloon dataset 
was designed to test TRN on a wide range of high-altitude data and occured in April of 2019. The New Shepard dataset was intended to 
test TRN on a high speed vehicle with a flight profile similar to that of a precision landing and occured in August of 2022.

\subsection{Balloon Flight}

\begin{figure}[hbt!]
    \centering
    \includegraphics[width=.4\textwidth]{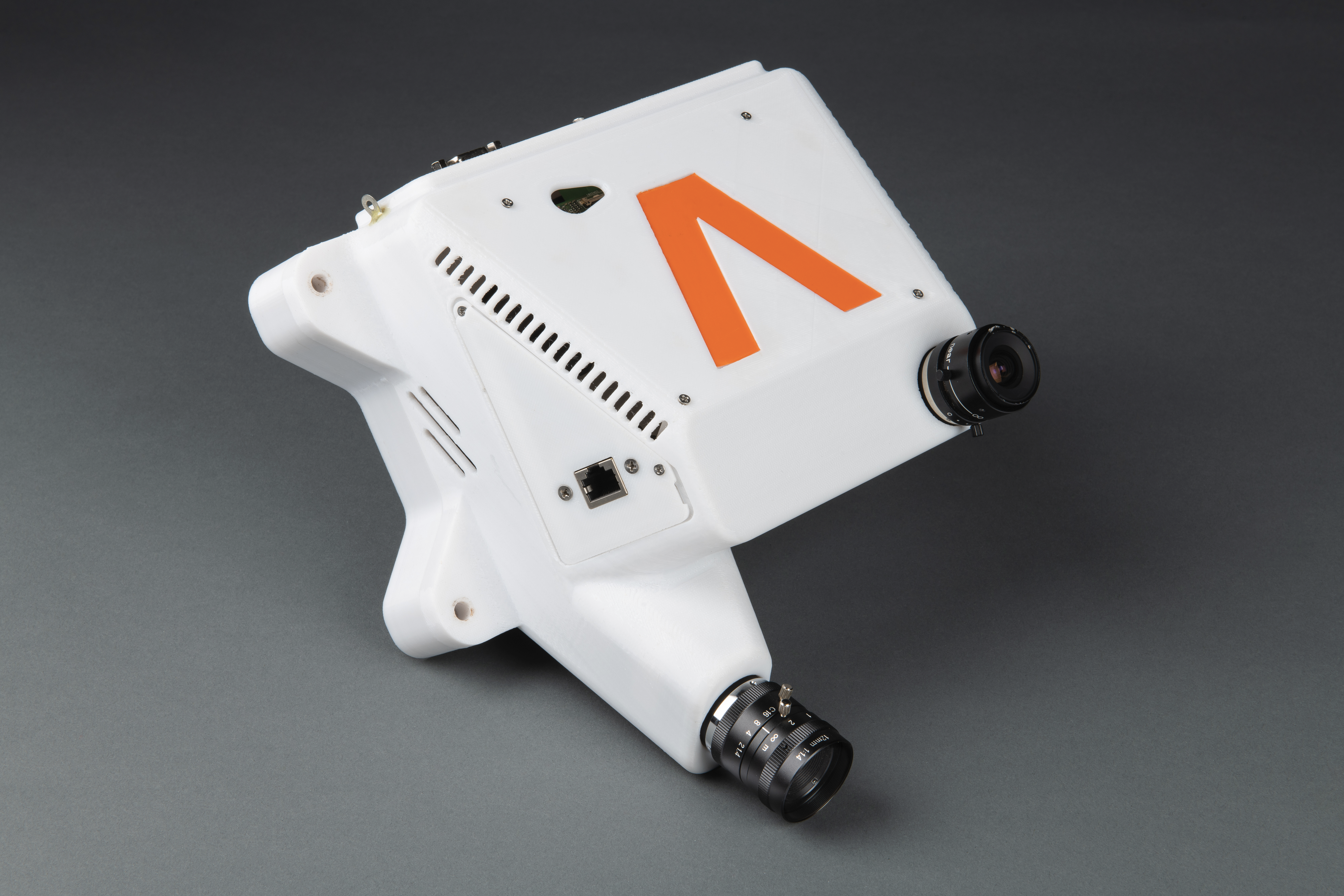}
    \caption{Draper Multi-Environment Navigator (DMEN) package: data collection package containing sideways and downward facing cameras, IMU, and logging computer.}
    \label{fig:hardware}
\end{figure}

We captured downward and sideways camera images along with data from a GPS and an inertial measurement unit (IMU) on board a World View 
Enterprises high-altitude balloon shown in \cref{fig:balloon_launch}, 
with data recorded up to an altitude of 33 km.
We used FLIR Blackfly S Color 3.2 MP cameras for both downward and sideways facing views using 12 mm EFL lens and 4.5 mm EFL lens, respectively. 
The field of view (FOV) for the downward and sideways camera with their respective lens is $32^{\circ}$ and $76^{\circ}$.
Both cameras, along with the IMU (Analog Devices ADIS16448) 
and data logging computer are self contained inside the Draper Multi-Environment Navigator (DMEN) package, shown in \cref{fig:hardware}. 
Both cameras generated images at 20 Hz with a resolution of $1024 \times 768$. The IMU logged data at 820 Hz. 

As mentioned in \cref{sec:intro}, some TRN applications ---such as 
planetary landing--- might prefer using a sideways-angled camera, while other applications 
---such as 
high-altitude drone flights--- may prefer a downward-facing camera. Therefore, we collect data from 
both a downward and sideways angled camera to allow for IBAL to be evaluated at both these camera 
angles. Some planetary landings may also desire a downward-facing camera since it allows the boresight of the camera 
to be normal to the surface during the terminal descent phase, 
such as was done for OSIRIS-REx \cite{Lorenz17ac-osirisrex}.

\subsection{Blue Origin New Shepard Flight}

We captured images from two sideways-angled cameras with 12.5 mm lens on opposite sides inside the New Shepard capsule which  
look out the capsule windows. Having two cameras was intended to allow us to study the effects of different cloud cover, terrain, and angle to the sun. 
We will refer to these cameras as camera 1 and camera 2. 
We additionally log IMU data from a Analog Devices ADIS16448, and 
telemetry from the capsule which served as ground truth for our experiment. Data was logged with a NUC mounted inside a payload locker in the capsule.
Both cameras generated images at 20 Hz with a resolution of $1024 \times 768$ and FOV of $31^{\circ}$. 
The IMU logged data at 820 Hz. The rocket reached speeds up to 880 km/h and an altitude of 8.5 km before an anomaly occurred during the NS-23 flight 
which triggered the capsule escape system.

\begin{figure}[hbt!]
    \centering
    \begin{subfigure}[t]{0.49\textwidth}
        \centering
        \includegraphics[width=\textwidth]{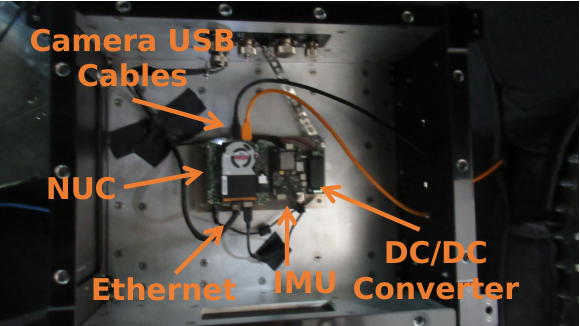}
    \end{subfigure}
    \hfil
    \begin{subfigure}[t]{0.49\textwidth}
        \centering
        \includegraphics[width=\textwidth]{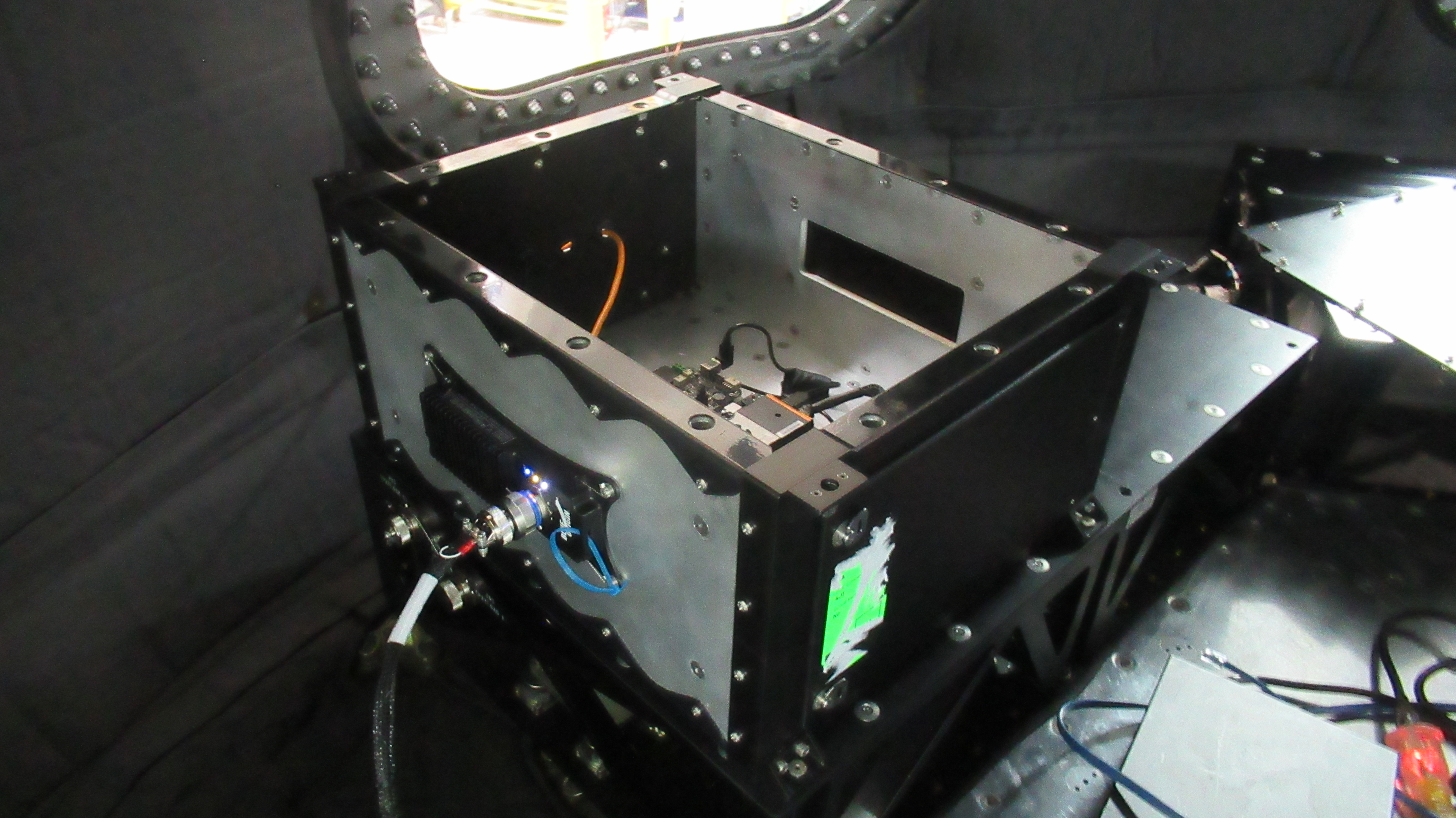}
    \end{subfigure}
    \caption{Payload locker inside the New Shepard capsule containing a NUC, IMU, and DC/DC Converter. Images courtesy of Blue Origin.}
    \label{fig:payload_blue}
\end{figure}

\Cref{fig:payload_blue} shows our payload locker containing the NUC, IMU, and a power converter which is 
mounted inside the New Shepard capsule. An ethernet cable and two USB cables transfer 
telemetry data from the capsule and data from the cameras to the NUC, respectively.

\begin{figure}[H]
    \centering
    \begin{subfigure}[t]{0.49\textwidth}
        \centering
        \includegraphics[width=\textwidth]{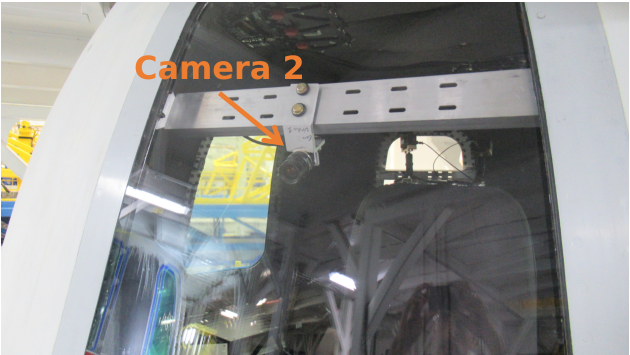}
        \caption{}
        \label{fig:cam_mount_a}
    \end{subfigure}
    \hfil
    \begin{subfigure}[t]{0.49\textwidth}
        \centering
        \includegraphics[width=\textwidth]{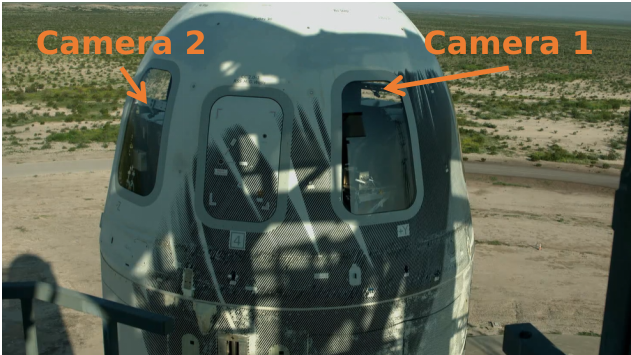}
        \caption{}
        \label{fig:cam_mount_b}
    \end{subfigure}
    \caption{Cameras 1 and 2 mounted inside the New Shepard capsule looking out the capsule windows. Images courtesy of Blue Origin.}
    \label{fig:cameras_in_window}
\end{figure}

\Cref{fig:cam_mount_a} shows camera 2 mounted inside the capsule with a sideways-angle and 
\cref{fig:cam_mount_b} shows the location of both cameras inside the capsule on opposite sides while New Shepard 
is on the launch pad. Both cameras are mounted at the same tilt angle such that they can view the terrain while not 
having their FOVs obstructed by components on the rocket. Additionally, a mounting angle was selected to reduce 
the effects of distortion caused by the windows, and to ensure the cameras did not come in direct 
contact with the windows.

Distortion effects from the windows were addressed by calibrating the instrinsic parameters 
of the camera while the camera was mounted in the capsule (i.e., a calibration board was positioned outside 
the capsule window). We used the Brown-Conrady model \cite{Brown66-brownConrady} which helps account for decentralized distortion caused by the window 
in addition to distortion from the camera lens. Further evaluation on the effects of distortion caused 
by the window of the capsule is left as a topic for future work.

\section{Terrain Relative Navigation Method}
\label{sec:method}

We use Draper's IBAL  software~\cite{Denver17aiaa-airdrop} to perform TRN 
for our datasets. A database of image templates is created in advance from satellite imagery and stored using known 
pixel correspondence with the world frame. Using satellite images and elevation maps from USGS~\cite{usgs}, we automatically select patches of interest 
from the satellite images and create a collection of templates that serve as 3D landmarks. For each camera image processed by IBAL, IBAL uses an initial guess of 
the camera pose to predict 
which templates from the database are in the field of view (FOV) of the camera using a projection from 
the image plane to an ellipsoidal model of the planet. The templates are then matched to the camera 
image using cross correlation. The resulting match locations are passed to a 3-point RANSAC \cite{Fischler81} (using a Perspective-Three-Point method as a minimal solver) to reject outliers. 
The output is a list of the inlier matches, their pixel location in the image, and their known location 
in the world frame that can be passed to a nonlinear estimator or fixed-lag smoother for tightly-coupled pose estimation. 
A secondary output of RANSAC is an absolute pose estimate found by using the Perspective-n-Point (PnP) 
algorithm on the set of inliers. 

Instead of a tightly-coupled approach, we will use a simpler method to evaluate performance on the balloon and New Shepard datasets. 
For the balloon dataset, we take the PnP absolute pose estimate directly from IBAL, 
 forward propagate it with the gyroscope measurements, and use it at the next time step as a pose guess for IBAL. 
 We do not use accelerometer data since 
in the image frame most scene changes for the balloon dataset 
over a short time span will be due to rotations. This
is due to the high altitude and hence large distance between the camera and the Earth's surface. Using the gyroscope to propagate the rotation also allows for
reduced computation since we are able to down-sample our camera data by a factor of 10 (2Hz image input to IBAL). 
Additionally, the gyro allows for robust handling of rapid motions of the balloon and images that have large obstruction
from cords which makes generating landmark matches unreliable. An ablation study on incorporating the gyroscope with IBAL is provided in \cref{sec:gyro_ablation}.
 Since the New Shepard capsule does not experience rapid rotations like the balloon, we did not find it 
necessary to use the gryoscope to forward propagate the pose estimate for the New Shepard dataset.

We propagate the rotation estimate of the vehicle, $q^{cam_T}_{ecef}$ (i.e., the orientation of the earth-centered, earth-fixed frame 
w.r.t. the camera frame at time $T$, represented as a unit quaternion), to the time of the next processed image ($T+1$) 
with the gyro using second order strapdown quaternion expansion \cite{Mckern68mit-transforms}.  
Using 3-axis gyro measurements $\bm{\theta}$ and their magnitude $\omega = \|\bm{\theta}\|$, we compute the orientation $q_{IMU_{t+1}}^{IMU_t}$ between gyro measurements 
using the following equation
\begin{equation}
    \label{eq:quat_gyro1}
    q_{IMU_{t+1}}^{IMU_t}= [1 - \frac{\omega^2 \Delta t_{IMU}^2}{8}, \frac{\bm{\theta}^T \Delta t_{IMU}}{2}]
\end{equation}
where $t+1$ and $t$ represent the time of consecutive IMU measurements occuring $\Delta t_{IMU}$ seconds apart.

Using the rotations $q_{IMU_{t+1}}^{IMU_t}$ between consecutive IMU timestamps, we 
can compute the relative rotation $q_{cam_{T+1}}^{cam_T}$ between the camera pose between consecutive images collected at time $T$ and $T+1$:
\begin{equation}
    \label{eq:quat_gyro2}
    q_{cam_{T+1}}^{cam_T} = \prod_{t = T}^{T+1} q_{IMU}^{cam} \otimes q_{IMU_{t+1}}^{IMU_t} \otimes (q_{IMU}^{cam})^{-1}
\end{equation}
where $\otimes$ is the quaternion product and $q_{IMU}^{cam}$ is the static transform from the IMU frame to the camera frame:

Finally, we can compute the rotation estimate $q^{cam_{T+1}}_{ecef}$ of the vehicle at time $T+1$:
\begin{equation}
    \label{eq:quat_gyro3}
    q^{cam_{T+1}}_{ecef} = (q_{cam_{T+1}}^{cam_T})^{-1} \otimes q^{cam_{T}}_{ecef}
\end{equation}

We use a simple yet effective logic for handling short segments in our datasets when PnP is unable to produce a reliable pose, which can be caused by image obstructions 
or blurry images caused by rapid vehicle motion. If PnP RANSAC selects a small set of inliers (i.e., less than 8) or if the pose is clearly infeasible (i.e., an altitude change between 
processed images greater than 450 m for the balloon dataset), we reject the 
pose estimate, keep forward propagating the pose using gyroscope data, and run IBAL with the next available image, ignoring the down-sampling rate. %

\section{Addressing Challenges of High-Altitude Images}

We apply simple and effective methods to address two common challenges we encountered with high-altitude images, namely determining the projection
to the ellipsoid when the camera views the horizon, and reducing the number of potential landmarks from the database that have a lower probability of 
generating good matches when there is a large number of landmarks in view of the camera. 

\begin{figure}[H]
        \centering
        \includegraphics[width=.3\textwidth]{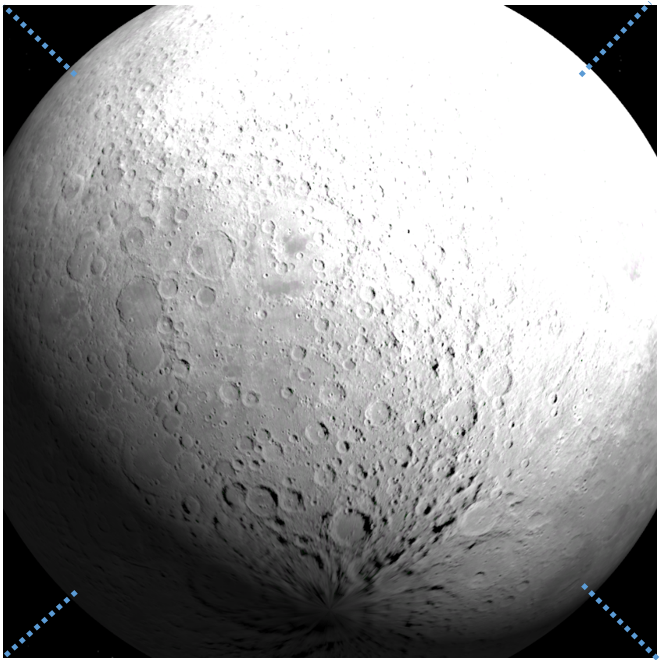}
        \caption{Example of our horizon detection method finding the horizon of an ellipsoidal body. Each corner point of the image 
        is incremented towards the opposite corner until the ellipsoid body is intersected.}
        \label{fig:works}
\end{figure}

When the horizon is in view of the camera, as is true for the higher altitude images from the sideways camera for the balloon dataset 
(\cref{fig:overview}), our 
baseline method of determining the camera's viewing bounds of the planet's surface is insufficient. Our baseline method is to use an 
initial estimate of the camera's pose to project each corner of the image to the ellipsoid model. From this, we can create a bounding box on the 
ellipsoid defined by a minimum and maximum latitude and longitude. However, this is ill-defined if at least one corner of the image falls 
above the horizon. 
To resolve this case, if the projection of a corner point does not intersect the ellipsoid we incrementally 
move the point (in the image space) towards the opposite corner of the image until it intersects the ellipsoid (\cref{fig:works}). This process is summarized in \cref{alg:horizon_detection}. 
This process is shown to be effective for our dataset, despite the fact that the approach could fail (see line 15 in~\cref{alg:horizon_detection}) when the projection of the ellipsoid does not intersect the main diagonals of the image (e.g., when the camera is too far away from Earth or has a large tilt angle).

\begin{algorithm}
    \caption{Horizon Detection} 
    \label{alg:horizon_detection}
    \small
    \begin{algorithmic}[1]
        \State \textbf{Inputs:} 
            \State \indent \indent P \Comment{estimate of camera projection matrix (containing intrinsic and extrinsic parameters)}
            \State \indent \indent $\pi_{WGS84}$  \Comment{projection of a pixel coordinate to a 3D point on the surface of the WGS84 model}
            \State \indent \indent $\delta x$ \Comment{amount to shift a point by in pixel space (default 10 pixels)}

        \State \textbf{Output:} $image\_corners$ \Comment{set of four pixel coordinates bounding image}
        
        \For{$x_{corner}, \in image\_corners$}
            \While{True}
                \State X $\gets \pi_{WGS84}(P, x_{corner})$
                \If{X intersects ellipsoid} 
                    \State break \Comment{found valid image boundary}
                \Else
                    \State increment $x_{corner}$ towards opposite corner by $\delta x$
                \EndIf
                \If{$x_{corner}$ outside image}
                    \State \textbf{return} error \Comment{failed to find horizon boundary}
                \EndIf
            \EndWhile
        \EndFor
        \State \textbf{return} $image\_corners$
    \end{algorithmic}
\end{algorithm}

\begin{figure}[H]
    \centering
    \begin{subfigure}[t]{0.46\textwidth}
        \centering
        \includegraphics[width=\textwidth]{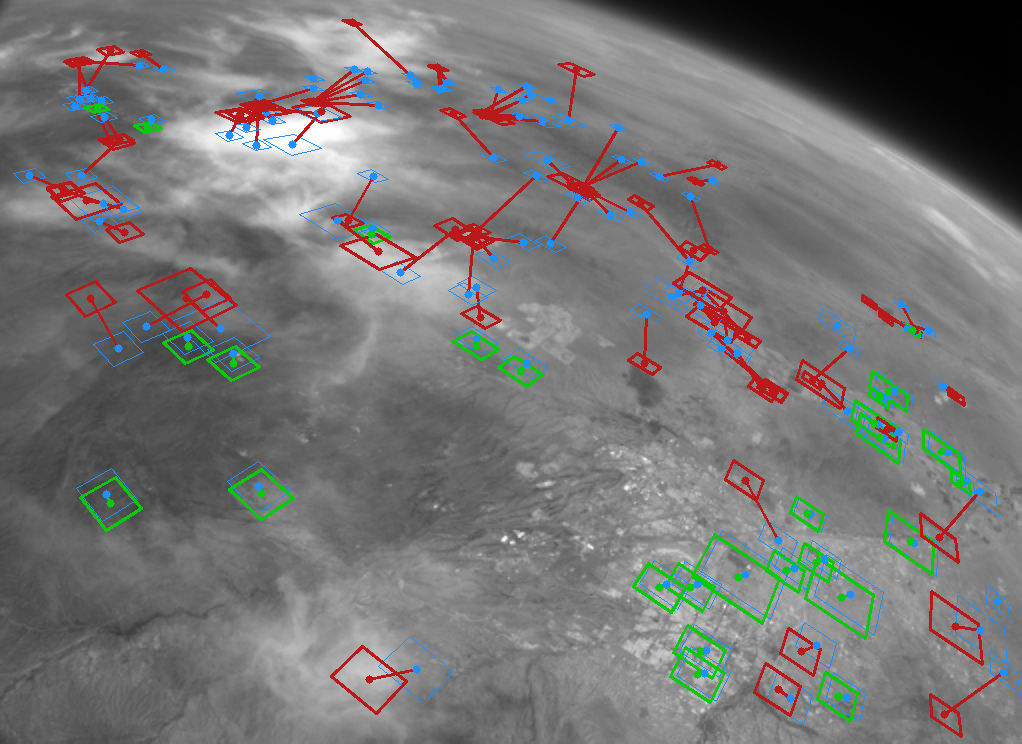}
        \caption{Higher concentration of outliers near the horizon without using landmark angle filter. Ratio of inliers to outliers: 0.3}
        \label{fig:landmark_angle_no_angle}
    \end{subfigure}
    \hfil
    \begin{subfigure}[t]{0.46\textwidth}
        \centering
        \includegraphics[width=\textwidth]{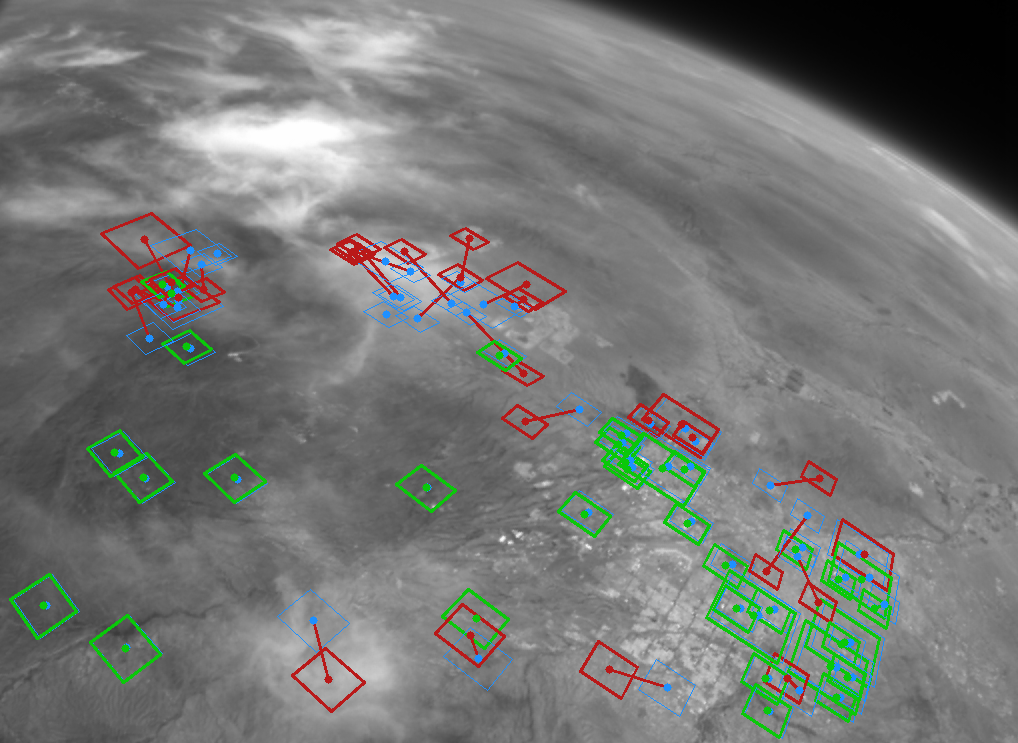}
        \caption{Higher concentration of inliers using landmark angle filter. Ratio of inliers to outliers: 1.3}
        \label{fig:landmark_angle_with_angle}
    \end{subfigure}
    \caption{Ablation study for \cref{alg:landmark_angle}, which filters regions of the image for landmark matching 
    based on the angle between the surface and the camera boresight. This leads to a higher ratio of inliers to outliers, reducing computation and improving accuracy. 
    Inliers matches are shown in green and outlier are shown in red. 
    Blue shows initial estimate of landmark location based on initial pose estimate before utilizing cross correlation. Images are from sideways camera 
    from balloon dataset.}
    \label{fig:angle_ablation}
\end{figure}

Since we select a maximum number of landmarks based on the landmarks in our satellite database that are in view of the camera, we need additional logic to
avoid the possibility of selecting landmarks that mostly fall near the horizon, since these are unlikely to lead to good matches. 
The ratio
of meters per pixels grows rapidly as we approach the horizon, and image matching becomes difficult or impossible near
the horizon line due to glare or heavy warping needed to match a shallow surface angle. Additionally, there 
is significant atmospheric distortion. 
Removing those landmarks helps avoid unnecessary computation and reduces the number of outliers we pass to RANSAC. Towards this goal,
 we set a maximum acceptable angle between the boresight of the 
camera and the surface normal of 
a landmark and reject landmarks that fail to meet this threshold. To increase the number of potential landmarks that 
meet our angle requirement, we filter out sections of the camera's FOV projection to the ellipsoid that are unlikely 
to produce landmarks that meet the angle threshold. 
This filtering method follows our prior method for intersecting the ellipsoid and uses 
similar logic. Starting at the first point near each image corner that views the ellipsoid, we find the surface normal by projecting 
from the image plane to the ellipsoid and move towards the oppostite corner of the image
until the angle requirement is met. This process is summarized in \cref{alg:landmark_angle} and a corresponding ablation 
is shown in \cref{fig:angle_ablation}. Notice that 
without \cref{alg:landmark_angle}, more landmarks are selected near the horizon (\cref{fig:landmark_angle_no_angle}) 
where template matching is more difficult resulting in more outliers. Using \cref{alg:landmark_angle} allows IBAL to target 
regions of the image with more distinguishable features for matching which results in a higher concentration of inliers 
(\cref{fig:landmark_angle_with_angle}).

\begin{algorithm}
    \small
    \caption{Landmark Angle Filter} 
    \label{alg:landmark_angle}
    \begin{algorithmic}[1]
        \State \textbf{Inputs:} 
            \State \indent \indent P \Comment{estimate of camera projection matrix (containing intrinsic and extrinsic parameters)}
            \State \indent \indent $\pi_{WGS84}$  \Comment{projection of a pixel coordinate to a 3D point on the surface of the WGS84 model}
            \State \indent \indent $\alpha_{max}$ \Comment{max acceptable angle between camera boresight and normal of a landmark}
            \State \indent \indent $\delta x$ \Comment{amount to shift a point by in pixel space (default 10 pixels)}
        
        \State \textbf{Output:} $image\_corners$ \Comment{set of four pixel coordinates bounding image}

        \State surface\_normal() $\gets$ function that finds normal vector at a point on the WGS84 model
        \State angle\_between() $\gets$ function that finds the angle between a camera boresight and a vector 
        \For{$x_{corner}, \in image\_corners$}
            \While{True}
                \State X $\gets \pi_{WGS84}(P, x_{corner})$
                \State $x_n \gets surface\_normal(X)$
                \State $\alpha \gets angle\_between(P, x_n)$
                \If{$\alpha \leq \alpha_{max}$}
                    \State break \Comment{found valid image bounary}
                \Else
                    \State increment $x_{corner}$ towards opposite corner by $\delta x$
                \EndIf
                \If{$x_{corner}$ outside image}
                    \State \textbf{return} error \Comment{failed to meet landmark angle requirement}
                \EndIf
            \EndWhile
        \EndFor
        \State \textbf{return} $image\_corners$
    \end{algorithmic}
\end{algorithm}

\section{Experiment Results}

\subsection{Balloon Flight}

\begin{figure}[hbt!]
    \centering
    \includegraphics[width=\textwidth, height=0.40\textheight]{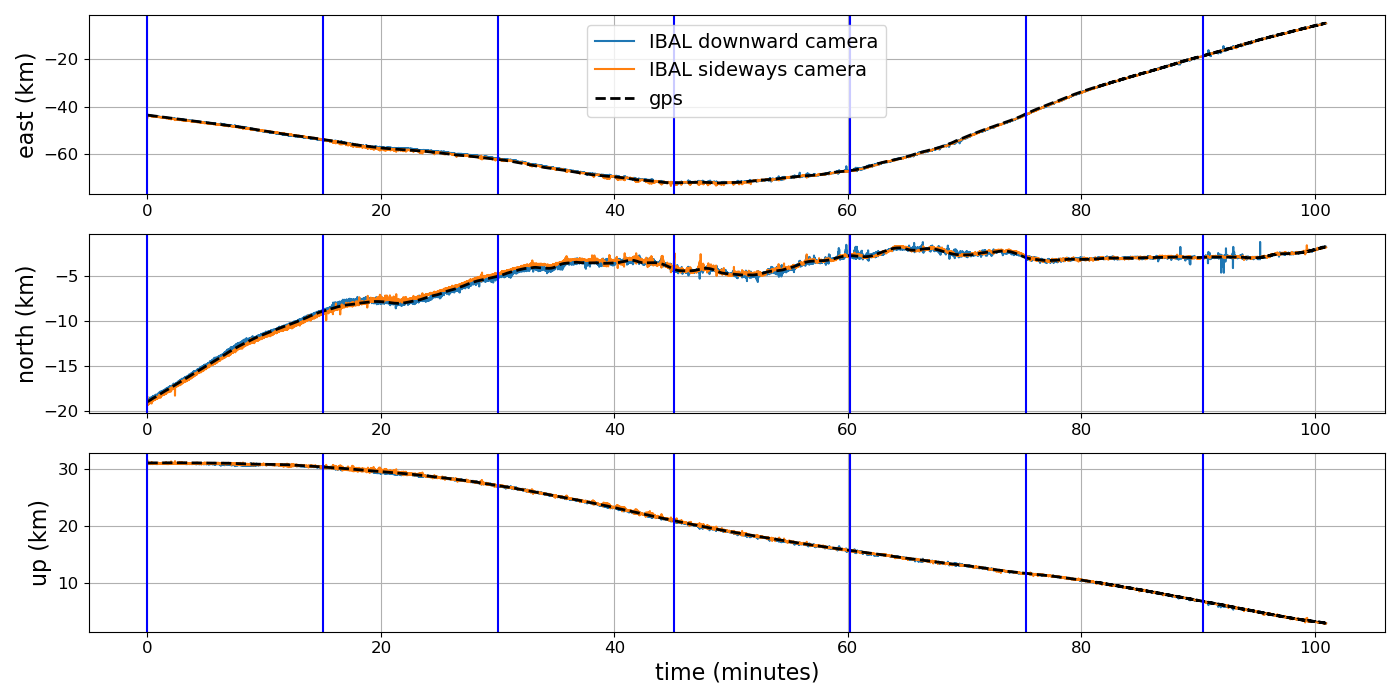}
    \caption{IBAL+gyro trajectory estimate vs. GPS for altitude range of 33 km to 4.5 km on balloon dataset. 
    Vertical lines show start of each new data segment.}
    \label{fig:all_trajectory}
\end{figure}

\begin{figure}[hbt!]
    \centering
    \includegraphics[width=\textwidth, height=0.40\textheight]{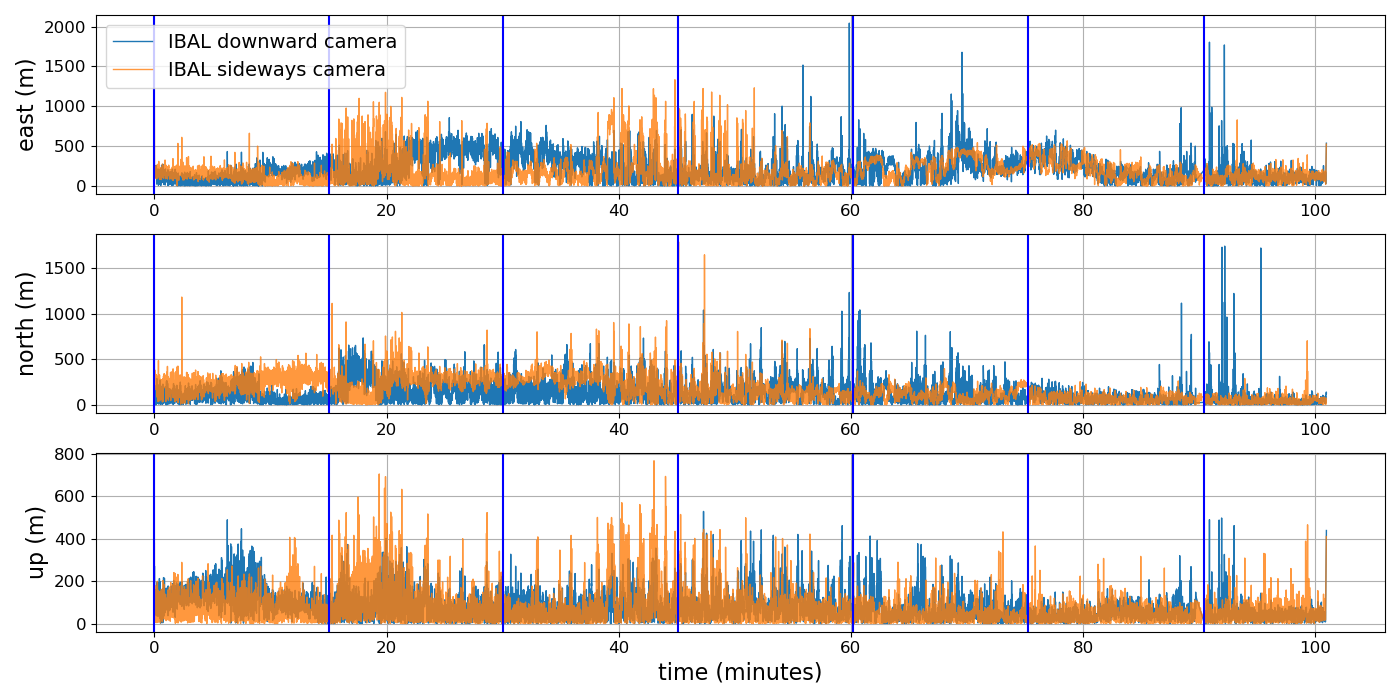}
    \caption{IBAL+gyro absolute position error for altitude range of 33 km to 4.5 km on balloon dataset. Vertical lines show start of each new data segment.}
    \label{fig:all_trajectory_error}
\end{figure}

\begin{figure}[H]
    \centering
    \includegraphics[width=\textwidth, height=0.37\textheight]{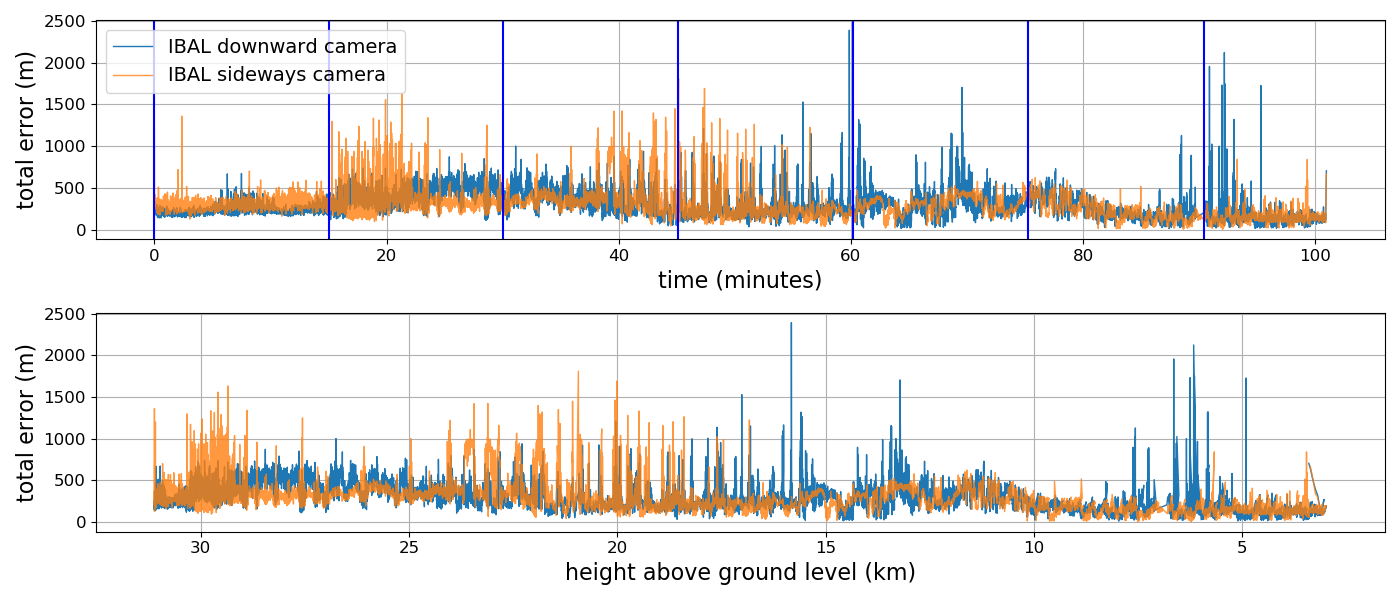}
    \caption{IBAL+gyro total trajectory error vs. time and vs. height above ground level on balloon dataset. Error tends to 
    show slight decrease in magnitude at lower altitudes. Vertical lines show start of each new data segment.}
    \label{fig:all_trajectory_total_error}
\end{figure}

We present results from running IBAL with both a sideways-tilted and downward-facing camera aided by gyroscope measurements on altitudes ranging from 33km to 4.5km. 
Note that we use the term altitude to mean height above the WGS84 ellipsoid. 
During this time, the system is descending under a parachute. 
We split our data into 
7 segments, each about 15 minutes long, and evaluate our estimated TRN position by comparing with GPS. We manually reseed IBAL at the start of each segment. 
Results are defined with respect to an East North Up (ENU) frame centered at the landing site of the balloon. 
\Cref{fig:all_trajectory} shows the ground truth trajectory from GPS compared to the trajectory estimates from IBAL with a downward and sideways 
facing camera. The corresponding plot of absolute position 
error is shown in \cref{fig:all_trajectory_error} for each of the East, North, and Up axes. 
IBAL is able to achieve an average position error along the up axis of 78 m and 66 m for the entire trajectory with the downward-facing and sideways-tilted camera, 
respectively, while the balloon travels almost 30 km in elevation.
IBAL achieves 207 m and 124 m of average position error for the east and north axis across the entire trajectory of the 
downward-facing camera, and likewise an average error of 177 m and 164 m along the east and north axis for the sideways camera 
while the balloon transverses well over 100 km laterally. 
\Cref{fig:all_trajectory_total_error} shows total absolute error (defined as the Euclidean distance between the estimate and the GPS position) with respect to flight time and with respect to height above ground level. 
Average absolute position error for the entire trajectory is 287 m and 284 m for the downward and sideways-tilted camera, respectively. 
Spikes in position estimates could be diminished 
using filtering methods such as coupling with an accelerometer or with visual odometry as mentioned in \cref{sec:method}. 
We run IBAL in real-time on a laptop with an Intel Xeon 10885M CPU. 
While IBAL is designed to run in real-time 
on flight hardware, we do not make showcasing run-time performance a focus of this paper.

\newpage

\newpage

\begin{figure}[H]
    \centering
    \begin{subfigure}[t]{1.0\textwidth}
        \centering
        \includegraphics[width=\textwidth, height=0.195\textheight]{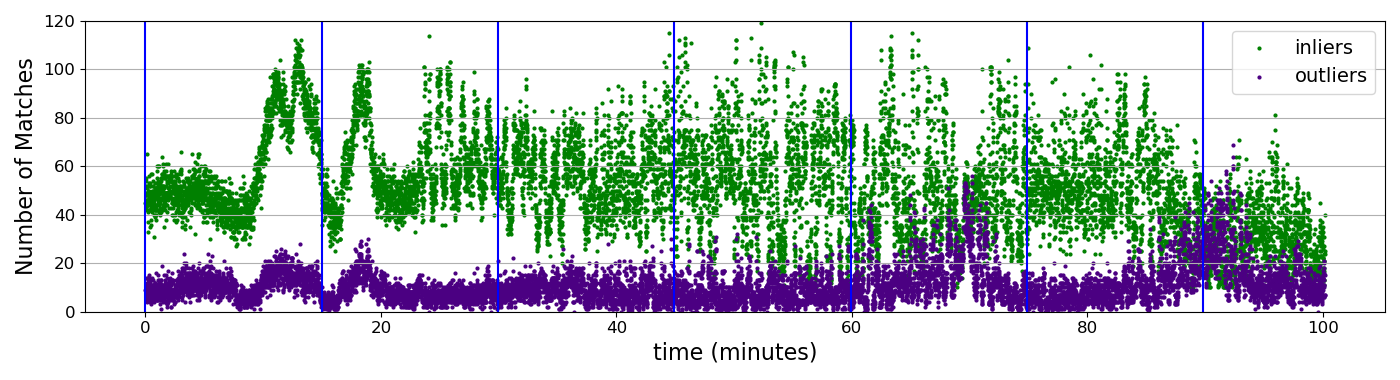}
        \caption{IBAL landmarking matching results for downward-facing camera}
        \label{fig:all_matches_down}
    \end{subfigure}

    \begin{subfigure}[t]{1.0\textwidth}
        \centering
        \includegraphics[width=\textwidth, height=0.195\textheight]{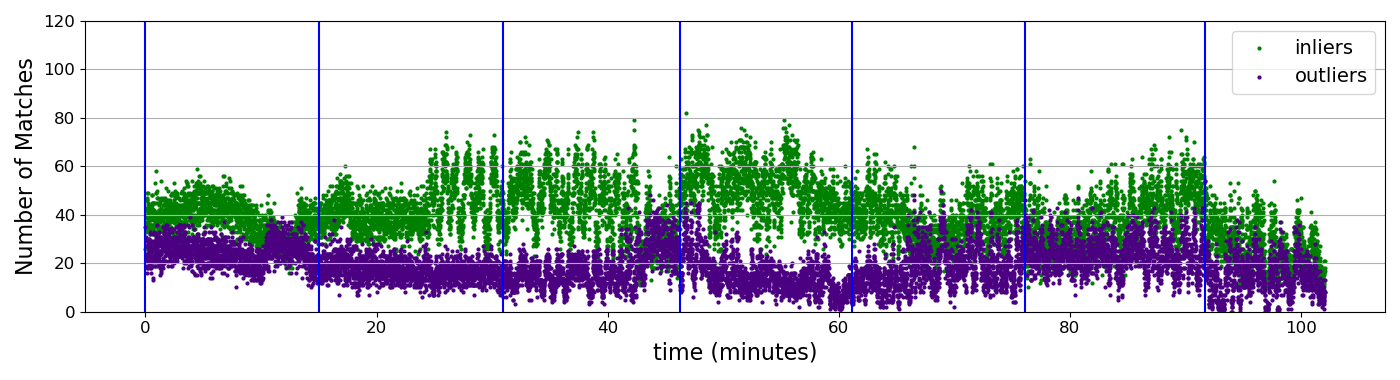}
        \caption{IBAL landmarking matching results for sideways-tilted camera}
        \label{fig:all_matches_side}
    \end{subfigure}
    \caption{IBAL+gyro number of inliers and outliers for sideways-tilted and downward-facing cameras on balloon dataset for altitude range of 33 km to 4.5 km 
    as determined by PnP and RANSAC. Vertical lines show start of each new data segment. The downward camera tends to have more matches than the sideways-tilted camera.}
\end{figure}

We also provide an analysis of the match correlation for both cameras for the entire balloon dataset. 
\Cref{fig:all_matches_down} and \cref{fig:all_matches_side} show number of inliers and outliers 
for the downward and sideways facing cameras. After estimating the location of a landmark in the image 
with cross correlation and peak finding, inliers and outliers are labeled using PnP and RANSAC. 
There are generally more inliers than outliers which shows the effectiveness of the correlation approach, and
that IBAL is able to perform well in the presence of outliers. 
We observe a greater number of inliers with the 
downward-facing camera than with the sideways-tilted camera.

\begin{figure}[H]
    \centering
    \begin{subfigure}[t]{0.48\textwidth}
        \centering
        \textbf{Downward Camera}\par\medskip
        \includegraphics[width=\textwidth, height=0.1\textheight]{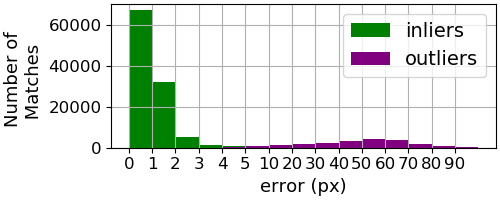}
        \caption{altitude range: 33 km to 32.5 km}
    \end{subfigure}
    \hfil
    \begin{subfigure}[t]{0.48\textwidth}
        \centering
        \textbf{Sideways Camera}\par\medskip
        \includegraphics[width=\textwidth, height=0.1\textheight]{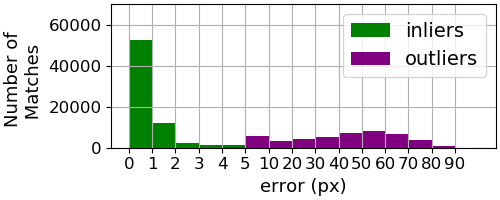}
        \caption{altitude range: 33 km to 32.5 km}
    \end{subfigure}

    \begin{subfigure}[t]{0.48\textwidth}
        \centering
        \includegraphics[width=\textwidth, height=0.1\textheight]{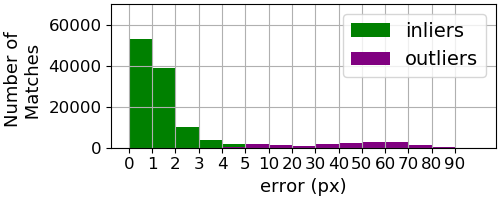}
        \caption{altitude range: 32.5 km to 29 km}
    \end{subfigure}
    \hfil
    \begin{subfigure}[t]{0.48\textwidth}
        \centering
        \includegraphics[width=\textwidth, height=0.1\textheight]{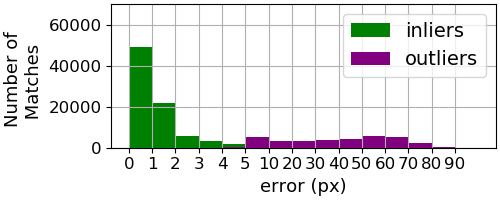}
        \caption{altitude range: 32.5 km to 29 km}
    \end{subfigure}

    \begin{subfigure}[t]{0.48\textwidth}
        \centering
        \includegraphics[width=\textwidth, height=0.1\textheight]{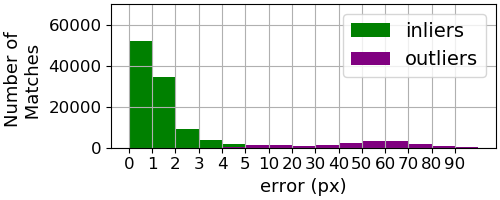}
        \caption{altitude range: 29 km to 23 km}
    \end{subfigure}
    \hfil
    \begin{subfigure}[t]{0.48\textwidth}
        \centering
        \includegraphics[width=\textwidth, height=0.1\textheight]{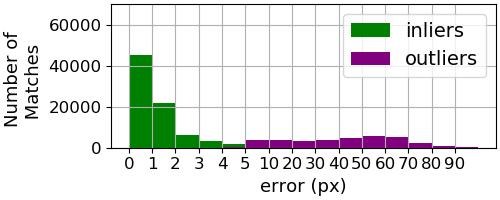}
        \caption{altitude range: 29 km to 23 km}
    \end{subfigure}

    \begin{subfigure}[t]{0.48\textwidth}
        \centering
        \includegraphics[width=\textwidth, height=0.1\textheight]{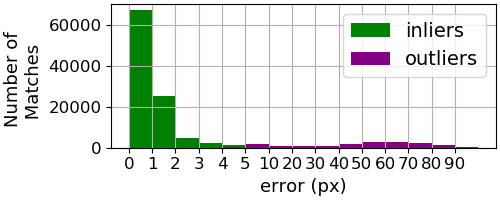}
        \caption{altitude range: 23 km to 18 km}
    \end{subfigure}
    \hfil
    \begin{subfigure}[t]{0.48\textwidth}
        \centering
        \includegraphics[width=\textwidth, height=0.1\textheight]{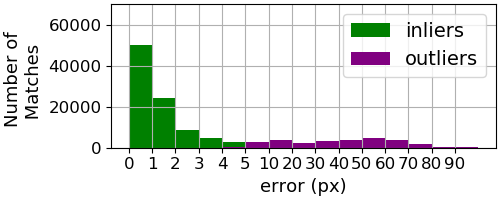}
        \caption{altitude range: 23 km to 18 km}
    \end{subfigure}

    \begin{subfigure}[t]{0.48\textwidth}
        \centering
        \includegraphics[width=\textwidth, height=0.1\textheight]{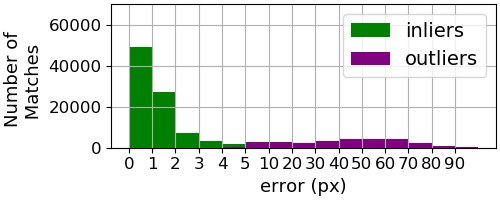}
        \caption{altitude range: 18 km to 14 km}
    \end{subfigure}
    \hfil
    \begin{subfigure}[t]{0.48\textwidth}
        \centering
        \includegraphics[width=\textwidth, height=0.1\textheight]{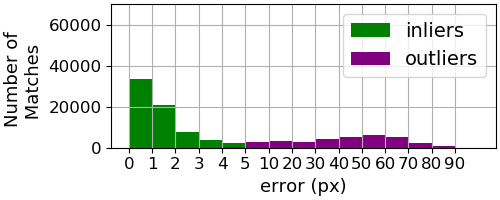}
        \caption{altitude range: 18 km to 14 km}
    \end{subfigure}

    \begin{subfigure}[t]{0.48\textwidth}
        \centering
        \includegraphics[width=\textwidth, height=0.1\textheight]{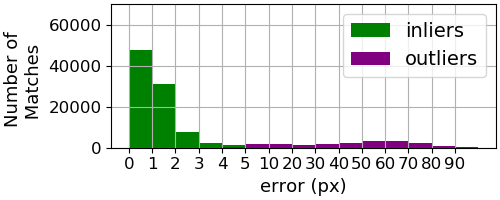}
        \caption{altitude range: 14 km to 9 km}
    \end{subfigure}
    \hfil
    \begin{subfigure}[t]{0.48\textwidth}
        \centering
        \includegraphics[width=\textwidth, height=0.1\textheight]{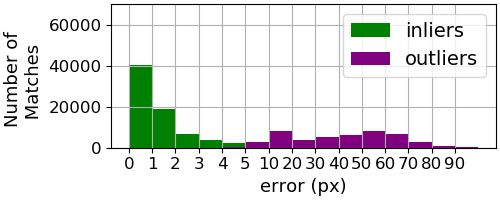}
        \caption{altitude range: 14 km to 9 km}
    \end{subfigure}

    \begin{subfigure}[t]{0.48\textwidth}
        \centering
        \includegraphics[width=\textwidth, height=0.1\textheight]{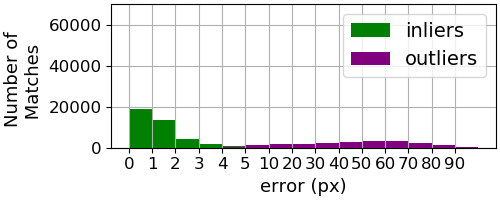}
        \caption{altitude range: 9 km to 4.5 km}
    \end{subfigure}
    \begin{subfigure}[t]{0.48\textwidth}
        \centering
        \includegraphics[width=\textwidth, height=0.1\textheight]{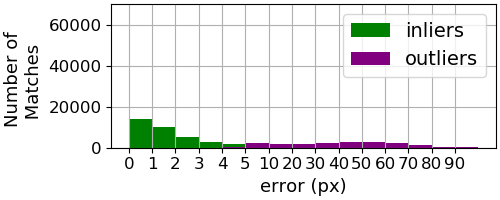}
        \caption{altitude range: 9 km to 4.5 km}
    \end{subfigure}

\caption{Inlier and outlier pixel error for each segment of balloon dataset. Error is the reprojection error determined by PnP and RANSAC. 
Left Column: downward camera, Right Column: sideways camera. 
Rows correspond to different altitude ranges.
}
\label{fig:histograms}
\end{figure}

Additionally, \cref{fig:histograms} shows a histogram of the amount of pixel error for the inliers and outliers 
determined by PnP and RANSAC for both the downward and sideways-tilted cameras. Inlier pixel error is distributed such that 
most inliers have between 0 and 1 pixel of error as determined by PnP and RANSAC which shows the effectiveness of IBAL's correlation approach. 
That there is an increase in the ratio of outliers to inliers at lower altitudes. This is due in part to shadows, lack of distinct texture on the ground, and 
regions with a sparse amount of landmarks in our database. 
Depending on mission requirements, this issue can be greatly reduced 
during the landmark database creation process such as by optimizing for landmark template size, ensuring sufficent landmark coverage at low altitudes 
for all phases of a flight, and by baking shadows into the database as was demonstrated in \cite{Smith22aiaa-blenderTRN}. 
However, for the purposes of the balloon experiment in this paper, we determined our database to be sufficient.

\begin{figure}[H]
    \centering
    \begin{subfigure}[t]{0.49\textwidth}
        \centering
        \includegraphics[width=\textwidth]{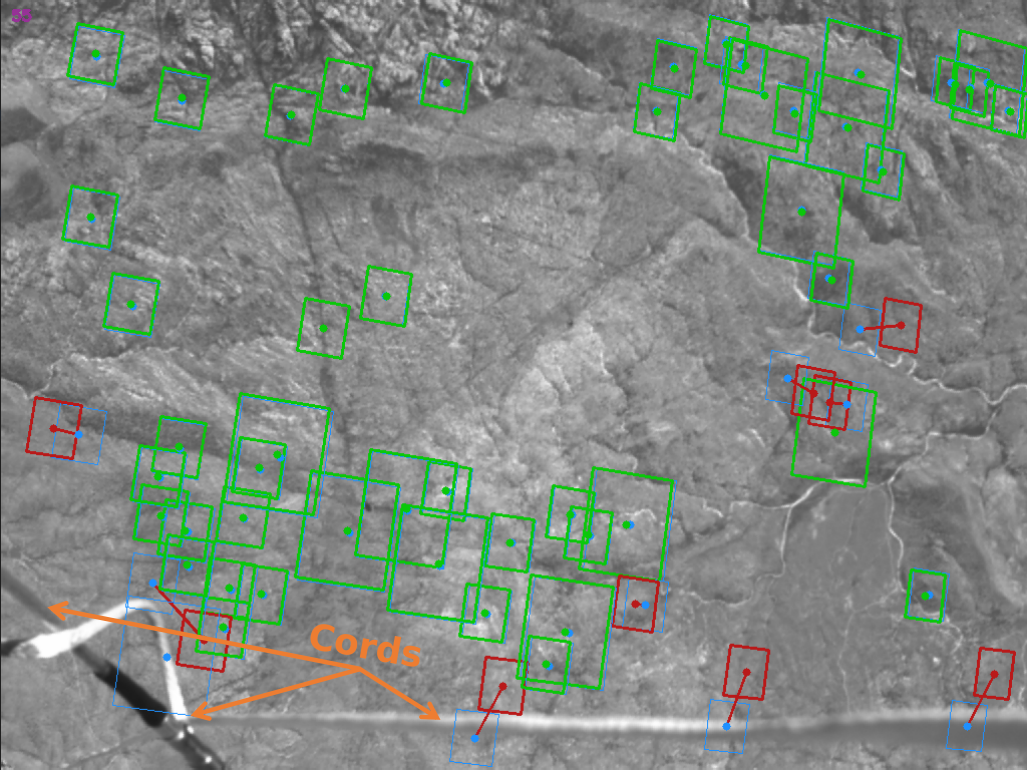}
        \caption{Downward Camera, altitude 13.5 km}
        \label{fig:down_match_135}
    \end{subfigure}
    \hfil
    \begin{subfigure}[t]{0.49\textwidth}
        \centering
        \includegraphics[width=\textwidth]{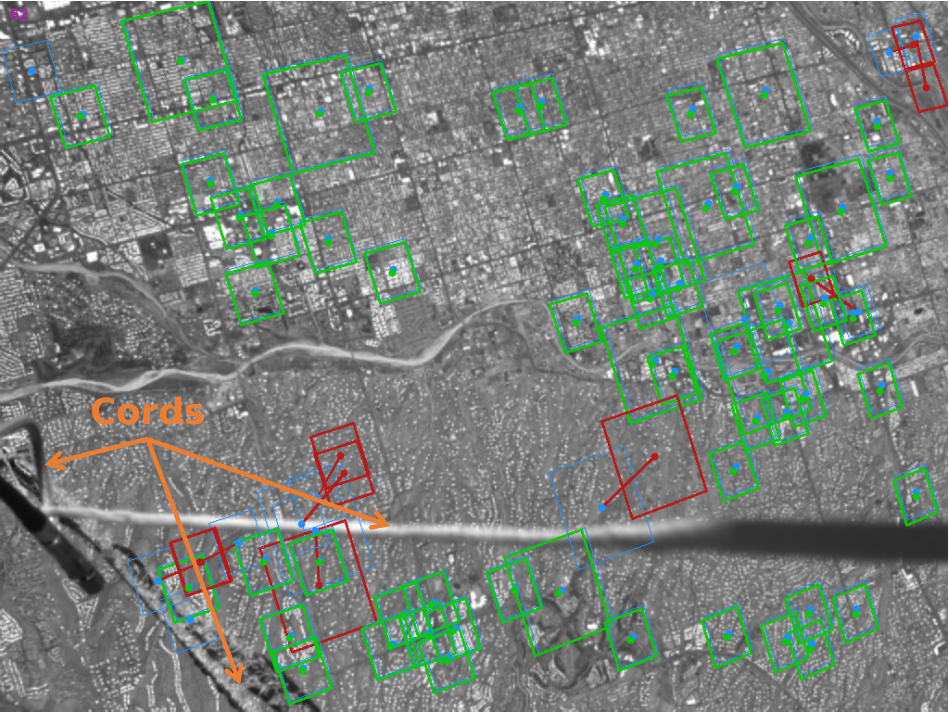}
        \caption{Downward Camera, altitude 23 km}
        \label{fig:down_match_23}
    \end{subfigure}

    \begin{subfigure}[t]{0.49\textwidth}
        \centering
        \includegraphics[width=\textwidth]{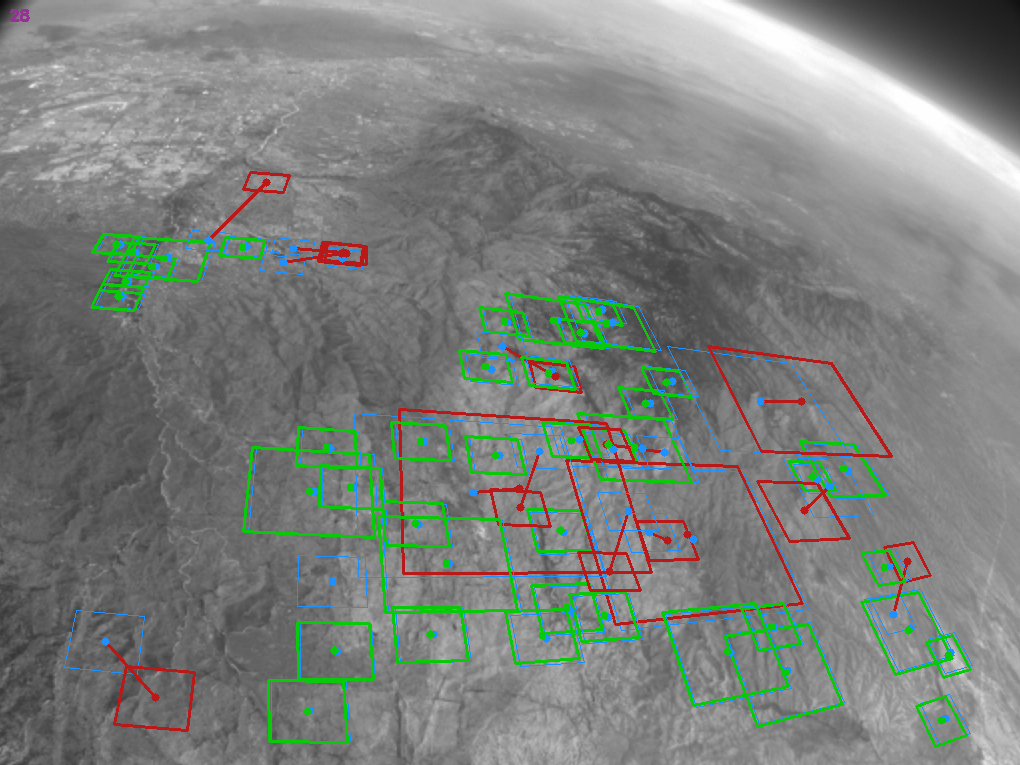}
        \caption{Sideways Camera, altitude 13.5 km}
        \label{fig:side_match_135}
    \end{subfigure}
    \hfil
    \begin{subfigure}[t]{0.49\textwidth}
        \centering
        \includegraphics[width=\textwidth]{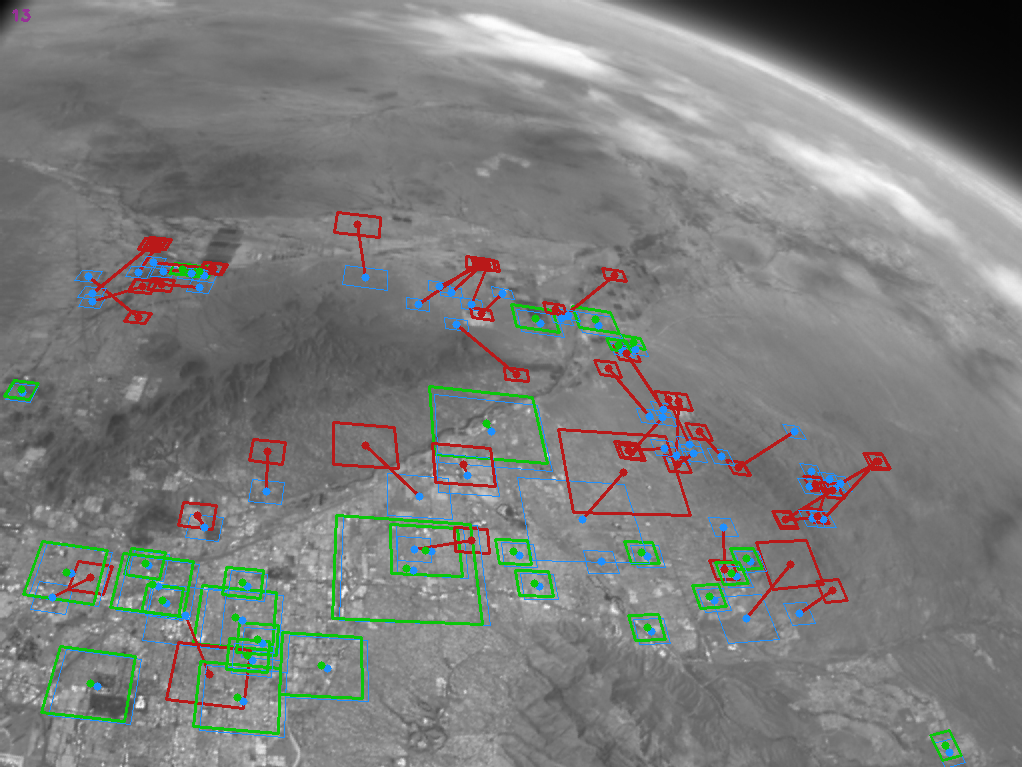}
        \caption{Sideways Camera, altitude 23 km}
        \label{fig:side_match_23}
    \end{subfigure}

    \caption{IBAL landmark match analysis on balloon dataset. Inliers matches are shown in green and outlier are shown in red. 
    Points in blue show initial estimate of landmark location based on initial pose estimate before utilizing cross correlation. Lines connect blue estimate 
    to calculated match location. Landmarks locations covered 
    by the cords are correctly rejected as outliers (top row).}
\end{figure}

Lastly, we provide visual examples of IBAL matches on a selected subset of frames from the downward and sideways facing cameras. 
\Cref{fig:down_match_135} shows landmark matches for the downward camera at 13.5 km with inliers shown in green and 
outliers shown in red. Blue dots show the inital estimate of the landmark locations in the image by using the pose estimated by IBAL's 
prior pose and the gyro before matching with cross correlation. 
\Cref{fig:down_match_23} shows matches for the downward camera at 23 km. 
Cords from the high-altitude balloon are partially in view, but incorrect matches caused by the cords are correctly 
rejected as outliers. \Cref{fig:side_match_135} and \cref{fig:side_match_23} show results for the sideways-tilted camera 
at 13.5 km and 23 km.

\subsection{Blue Origin New Shepard Flight}

We present results from running IBAL with two cameras (referred to as camera 1 and camera 2) mounted inside the Blue Origin New Shepard capsule. 
We only show results up to an altitude of approximately 8.5 km 
since there was an anomaly that occurred during flight NS-23 which triggered the capsule escape system. 
Nevertheless, we are still able to show IBAL working while the rocket achieves 
nominal speeds up to 880 km/h (550 mph). We seed the initial input image to IBAL using telemetry from New Shepard and then use the previous IBAL pose estimate 
as the initial pose guess for the next timestep. Unlike the balloon experiment, we do not incorporate the gyroscope measurement to forward propagate the 
pose estimate since the capsule does not experience significant rotations during its ascent.

\begin{figure}[hbt!]
    \centering
    \includegraphics[width=1.0\textwidth]{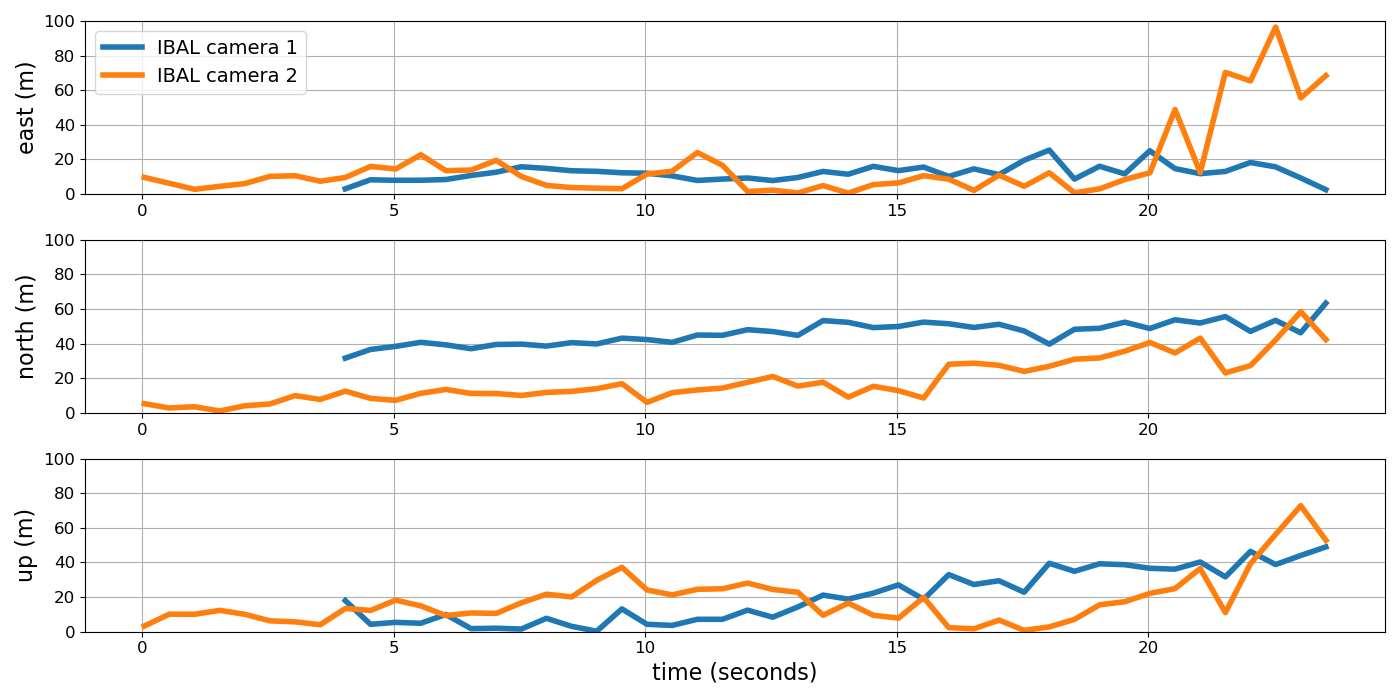}
    \caption{IBAL absolute position error on New Shepard dataset: altitude range of 3.5 km to 8.5 km.}
    \label{fig:blue_error}
\end{figure}

\begin{figure}[hbt!]
    \centering
    \includegraphics[width=1.0\textwidth]{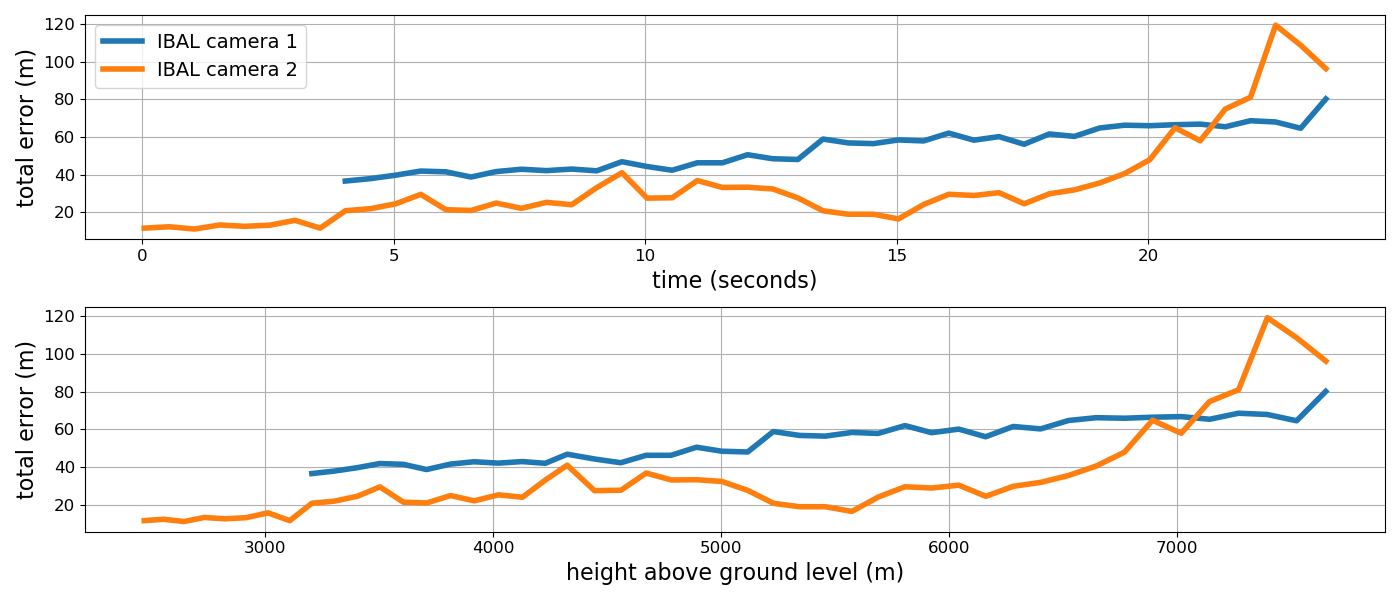}
    \caption{IBAL total trajectory error vs. time and height above ground level on New Shepard dataset. Total error is less than 120 m while reaching 
    speeds up to 880 km/h and a peak altitude of 8.5 km.}
    \label{fig:blue_total_error}
\end{figure}

We show a similar series of analysis of trajectory error and landmark matches as was presented for the high-altitude balloon experiment. 
Results are defined with respect to a ENU frame centered at the launch pad. 
\Cref{fig:blue_error} shows absolute error for each of the East, North, and Up axes by comparing the position estimate of IBAL with GPS. 
\Cref{fig:blue_total_error} shows total absolute error with respect to flight time and with respect to height above ground level. IBAL's total position 
error estimate is below 120 m for the duration of the dataset, and that error with camera 2 is as low as 10 m when the rocket is at an altitude of 3.5 km. 
Average absolute position error for the entire trajectory is 54 m and 34 m for camera 1 and camera 2, respectively. Both cameras show similar performance with IBAL, and 
slight differences in performance can be explained by the cameras being located on opposite sides of the capsule (and thus viewing different terrain) 
and by potential unaccounted distortion effects in the camera calibration.

\begin{figure}[H]
    \centering
    \begin{subfigure}[t]{0.46\textwidth}
        \centering
        \includegraphics[width=\textwidth]{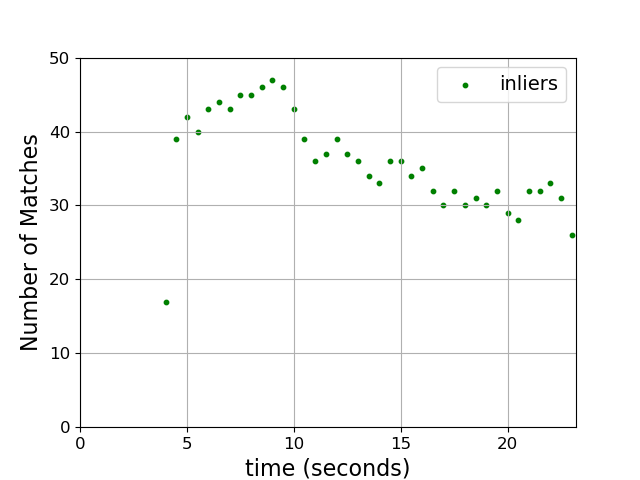}
        \caption{IBAL landmarking matching results for camera 1}
        \label{fig:blue_inliers_outliers_1}
    \end{subfigure}
    \hfil
    \begin{subfigure}[t]{0.46\textwidth}
        \centering
        \includegraphics[width=\textwidth]{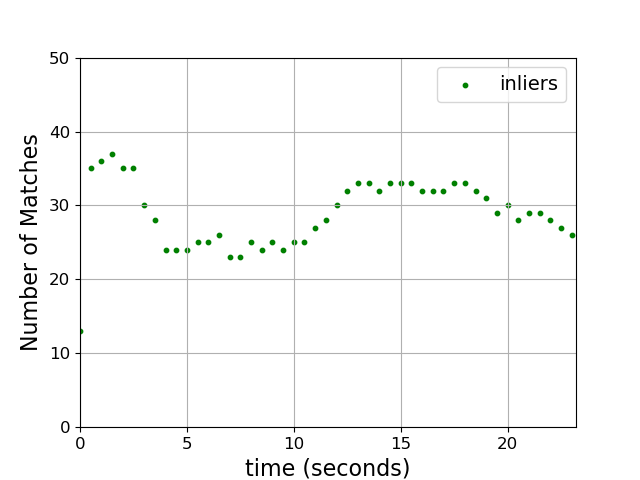}
        \caption{IBAL landmarking matching results for camera 2}
        \label{fig:blue_inliers_outliers_2}
    \end{subfigure}
    \caption{IBAL number of inliers and outliers for cameras 1 and 2 on New Shepard dataset as determined by PnP and RANSAC. 
    The data corresponds to an altitude range between 3.5 km and 8.5 km.}
\end{figure}

\begin{figure}[H]
    \centering
    \begin{subfigure}[t]{0.45\textwidth}
        \centering
        \includegraphics[width=\textwidth, height=0.1\textheight]{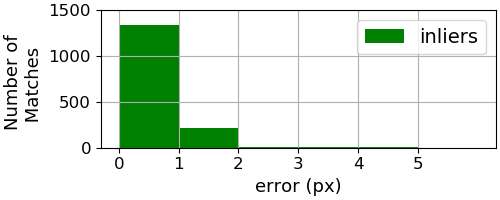}
        \caption{Camera 1}
    \end{subfigure}
    \hfil
    \begin{subfigure}[t]{0.45\textwidth}
        \centering
        \includegraphics[width=\textwidth, height=0.1\textheight]{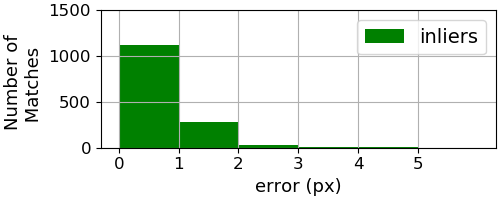}
        \caption{Camera 2}
    \end{subfigure}
    \caption{Inlier pixel error distribution for Cameras 1 and 2 on New Shepard dataset. 
    }
    \label{fig:blue_histogram}
\end{figure}

We also provide  an analysis of match correlation for both cameras. Since each processed frame only had at most 2 matches identified as outliers 
by PnP and RANSAC, we do not include match analysis for outliers in our results. 
\cref{fig:blue_inliers_outliers_1} and \cref{fig:blue_inliers_outliers_2}
show number of inliers for both cameras. 
\cref{fig:blue_histogram} shows a histogram of the amount of pixel error for the inliers determined by PnP
RANSAC for both cameras. Similarly to the results from the balloon flight, pixel error for a majority of the inliers is less than two pixels.

\begin{figure}[H]
    \centering
    \begin{subfigure}[t]{0.48\textwidth}
        \centering
        \includegraphics[width=\textwidth]{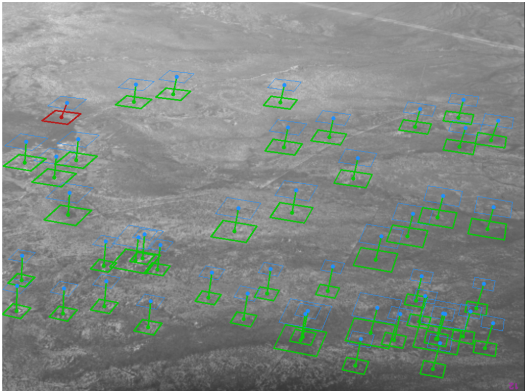}
        \caption{IBAL inlier and outlier matches for camera 1 on New Shepard dataset at an altitude of 6.4 km}
        \label{fig:blue_match_visualize_1}
    \end{subfigure}
    \hfil
    \begin{subfigure}[t]{0.476\textwidth}
        \centering
        \includegraphics[width=\textwidth]{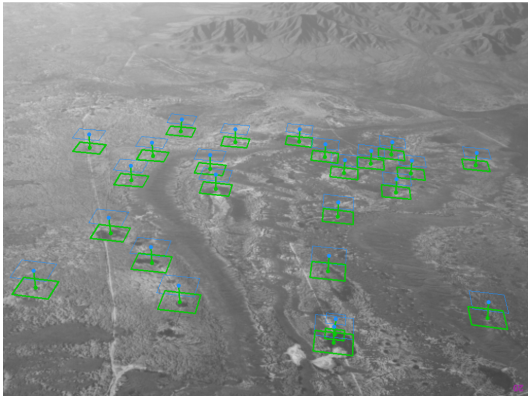}
        \caption{IBAL inlier and outlier matches for camera 2 on New Shepard dataset at an altitude of 6.4 km}
        \label{fig:blue_match_visualize_2}
    \end{subfigure}
    \caption{IBAL inlier and outlier matches for cameras 1 and 2 on New Shepard dataset. Inliers matches are shown in green and outlier are shown in red. 
    Blue shows initial estimate of landmark location based on initial pose estimate before utilizing cross correlation. Lines connect blue estimate 
    to calculated match location. Images have been rotated by 
    180 $^{\circ}$ for visual appeal.}
    \label{fig:blue_match_visualize}
\end{figure}

We provide visual examples of IBAL matches on a frame from both cameras in \cref{fig:blue_match_visualize}. 
Matches labeled as inliers are shown in green, while outliers are shown in red. There is only one outlier present in the processed image from 
camera 1 (\cref{fig:blue_match_visualize_1}) and no outliers in the image from camera 2 (\cref{fig:blue_match_visualize_2}).

\begin{figure}[H]
    \centering
    \includegraphics[width=0.46\textwidth]{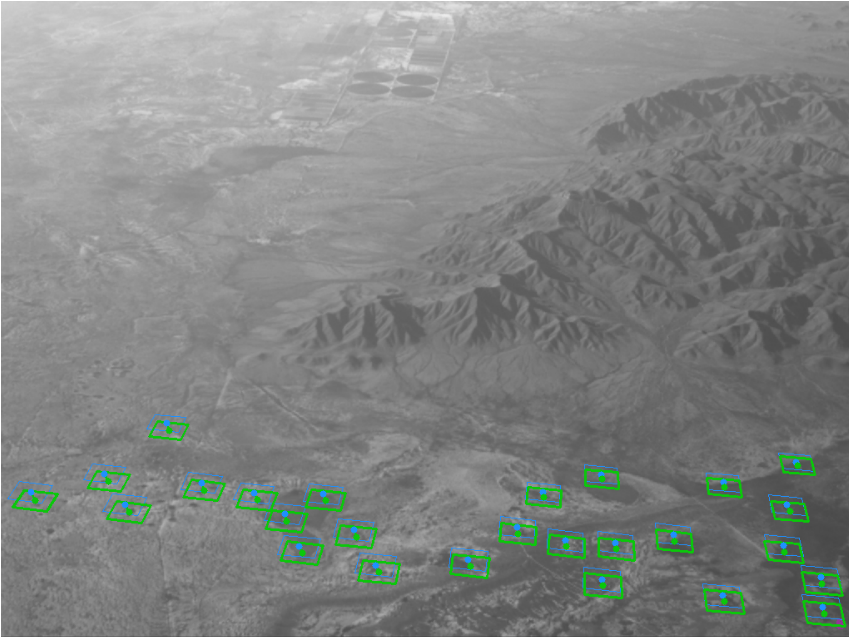}
    \caption{IBAL Camera 2 viewing a mountain range on New Shepard dataset. Inliers matches are shown in green. 
    Blue shows initial estimate of landmark location based on initial pose estimate before utilizing cross correlation. Lines connect blue estimate to calculated match location. 
    Image has been rotated by 
    180 $^{\circ}$ for visual appeal.}
    \label{fig:mountain}
\end{figure}

Lastly, we remark on one difficulty of the New Shepard dataset.
A mountain range is in view of camera 2 which makes landmark matching more difficult near the latter portion of the dataset as the mountain comes into the camera's 
FOV (\cref{fig:mountain}). This is due to the presence of shadows in the mountain that may not be consistent with shadows present in the 
time of day the database imagery was collected. Additionally, the 2D-2D homography assumption which we use to warp landmark templates into the image for 
correlation begins to break down when 3D structures 
such as mountains are viewed from low altitudes. Work with database creation such as \cite{Smith22aiaa-blenderTRN} along with advances in IBAL 
not mentioned in the paper can be used to reduce these issue for low altitude navigation over mountains.

\section{Gyroscope Incorporation Ablation Study}
\label{sec:gyro_ablation}

We provide an ablation study of forward propagating the IBAL pose estimate with a gyroscope for the high-altitude balloon dataset 
as mentioned in \cref{sec:method}. The benefits of incorporating the gyroscope data is two-fold. Firstly, since the balloon experiences 
rapid rotations, in some cases exceeding $20^\circ$ per second, the gyro provides a more accurate initial guess of the balloon's pose for IBAL, which 
reduces the frequency at which images must be to used to estimate the pose, hence reducing computation. Additionally, if landmark match quality is temporarily insufficient 
(typically on the order of 1 to 3 seconds) for PnP and RANSAC, which can be caused for example by significant obstruction by the cords below the balloon, the gyro allows the pose estimate to be carried over until good landmark matches can be found.

\Cref{table:gyro_ablation} shows the benefits of using the gyro with our balloon dataset. Using the downward-facing camera, we show the percentage 
of each of the seven data segments IBAL is able to successfully complete with and without incorporating the gyroscope. We also test on two different rates of 
image processing, noting that while one could partially compensate the lack of gyroscope measurements by increasing the rate of image processing,  that strategy is only effective at high altitudes in our dataset.

\begin{table}[H]
    \begin{tabular}{ | l | l | l | l | l | l | l | l |}
        \hline
        & 33-32.5 km & 32.5-29 km & 29-23 km & 23-18 km & 18-14 km & 14-9 km & 9-4.5 km\\ \hline
        4 Hz w/ gyro & 100 & 100 & 100 & 100 & 100 & 100 & 100  \\ \hline
        2 Hz w/ gyro & 100 & 100 & 100 & 100 & 100 & 100 & 100  \\ \hline
        4 Hz w/o gyro & 100 & 100 & 96 & 3 & 3 & 1 & 1  \\ \hline
        2 Hz w/o gyro & 100 & 100 & 63 & 0 & 0 & 1 & 1  \\ 
        \hline
    \end{tabular}
    \caption{Ablation study showing the benefit of incorporating gyroscope measurements with IBAL on each of the seven altitude segments of the balloon dataset 
    for different rates of image processing. 
    Results show the percent of each dataset segment IBAL successfully processes using images from the downward camera.}
    \label{table:gyro_ablation}
\end{table} %

\section{Conclusion}

This paper reports on the performance of a vision-based terrain relative navigation method on data ranging from 4.5 km to 33 km on a high-altitude 
balloon dataset and on data collected onboard Blue Origin's New Shepard rocket. We evaluate 
performance 
of both a sideways-tilted and downward-facing camera for the balloon dataset and two sideways-tilted 
cameras on the New Shepard dataset. We observe less than 290 meters of 
average position error on the balloon data over a trajectory of 150 kilometers and 
with the presence of rapid motions and dynamic obstructions in the field of view of the camera. Additionally, we report less than 55 m of 
average position error on the 
New Shepard dataset while reaching an altitude of 8.5 km and a max nominal speed of 880 km/h. As future work, we plan to fly again onboard the New Shepard 
rocket and capture camera data from ground level to an altitude of over 100 km. 
\section*{Acknowledgments}
We would like to gratefully acknowledge Andrew Olguin, Carlos Cruz, Alanna Ferri, Laura Henderson, and
everyone else at Draper who supported IBAL and data collection for the balloon flight and New Shepard flight. This
work was authored by employees of The Charles Stark Draper Laboratory, Inc. under Contract No. 80NSSC21K0348
with the National Aeronautics and Space Administration. The United States Government retains and the publisher, by
accepting the article for publication, acknowledges that the United States Government retains a non-exclusive, paid-up,
irrevocable, worldwide license to reproduce, prepare derivative works, distribute copies to the public, and perform
publicly and display publicly, or allow others to do so, for United States Government purposes. All other rights are
reserved by the copyright owner.

\end{document}